\documentclass[letterpaper, 10 pt, journal, twoside, pdftex]{IEEEtran}
\usepackage{cite}
\usepackage{hyperref}
\ifCLASSINFOpdf
  \usepackage[pdftex]{graphicx}
\else
  \usepackage[dvips]{graphicx}
\fi
\usepackage{amsmath}
\usepackage{booktabs}
\begin{document}
\title{Self-Supervised Learning to Fly using Efficient Semantic Segmentation and Metric Depth Estimation for Low-Cost Autonomous UAVs}
\author{Sebastian Mocanu$^{1}$, Emil Slusanschi$^{1}$, Marius Leordeanu$^{1, 2, 3}$%
\thanks{Sebastian Mocanu and Emil Slusanschi are with the $^{1}$National University of Science and Technology Politehnica Bucharest, 060042 Bucharest, Romania.
        (email: { \tt\footnotesize \href{mailto:sebastian.mocanu@upb.ro}{sebastian.mocanu@upb.ro}; \href{mailto:emil.slusanschi@cs.pub.ro}{emil.slusanschi@cs.pub.ro}})}%
\thanks{Marius Leordeanu is with the $^{1}$National University of Science and Technology Politehnica Bucharest, 060042 Bucharest, Romania, with the $^{2}$Institute of Mathematics "Simion Stoilow" of the Romanian Academy, 010702 Bucharest, Romania, and with $^{3}$NORCE Norwegian Research Center, Norway. 
        (email: {\tt\footnotesize \href{mailto:leordeanu@gmail.com}{leordeanu@gmail.com}})}%
}

\markboth{arXiv preprint.}
{Mocanu \MakeLowercase{\textit{et al.}}: Low-Cost Vision-Based Control Autonomous UAVs} 

\maketitle

\begin{abstract}
This paper presents a vision-only autonomous flight system for small UAVs operating in controlled indoor environments. The system combines semantic segmentation with monocular depth estimation to enable obstacle avoidance, scene exploration, and autonomous safe landing operations without requiring GPS or expensive sensors such as LiDAR. A key innovation is an adaptive scale factor algorithm that converts non-metric monocular depth predictions into accurate metric distance measurements by leveraging semantic ground plane detection and camera intrinsic parameters, achieving a mean distance error of 14.4 cm. The approach uses a knowledge distillation framework where a color-based Support Vector Machine (SVM) teacher generates training data for a lightweight U-Net student network (1.6M parameters) capable of real-time semantic segmentation. For more complex environments, the SVM teacher can be replaced with a state-of-the-art segmentation model. Testing was conducted in a controlled 5x4 meter laboratory environment with eight cardboard obstacles simulating urban structures. Extensive validation across 30 flight tests in a real-world environment and 100 flight tests in a digital-twin environment demonstrates that the combined segmentation and depth approach increases the distance traveled during surveillance and reduces mission time while maintaining 100\% success rates. The system is further optimized through end-to-end learning, where a compact student neural network learns complete flight policies from demonstration data generated by our best-performing method, achieving an 87.5\% autonomous mission success rate. This work advances practical vision-based drone navigation in structured environments, demonstrating solutions for metric depth estimation and computational efficiency challenges that enable deployment on resource-constrained platforms. 
\end{abstract}


\IEEEpeerreviewmaketitle

\section{Introduction}
\label{sec:introduction} 

\begin{figure}
    \centering
    \includegraphics[width=\linewidth]{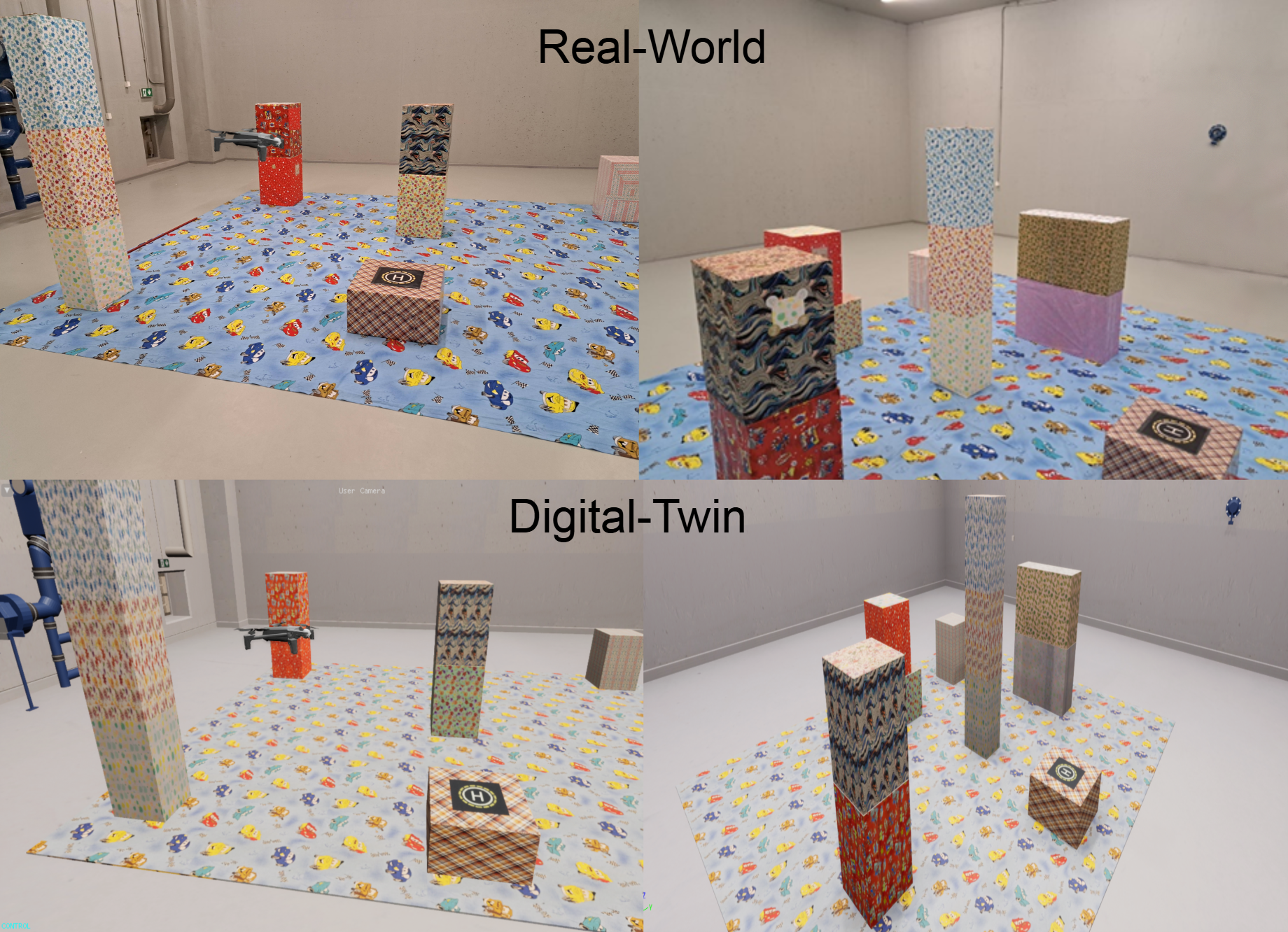}
    \caption{Comparison between real-world laboratory environment (top) and its corresponding digital-twin simulation environment (bottom). The digital-twin accurately replicates the physical environment geometry, object placement, textures and spatial relationships, enabling parallel experimentation and validation between physical and simulated domains.}
    \label{fig:dataset-imgs}
\end{figure}

Our proposed system operates at multiple levels. The primary level focuses on developing a system capable of recognizing semantic classes, which is essential for identifying obstacles, determining safe paths, and locating landing areas \cite{kakaletsis2021computer, a17040139}. While our initial implementation demonstrated efficient obstacle detection and base recognition, it was limited by the absence of depth information. To overcome this limitation, we introduced a multitask approach that integrates depth prediction to enhance obstacle avoidance and improve flight speed. This constitutes the second level of our system. However, most depth prediction systems are non-metric, which led us to develop an innovative algorithm to convert these predictions into accurate, metric depth maps by leveraging semantic segmentation data. By integrating semantic segmentation with depth estimation, we aim to create a comprehensive understanding of the drone's environment for safer and more efficient navigation \cite{8299437}. With the ground plane detected and the flight altitude estimated, we successfully scaled the depth measurements, enabling precise distance calculations for creating a safe flight corridor. This innovation significantly improved the UAV's ability to navigate efficiently while maintaining a sufficient safety margin.

By combining semantic understanding with monocular depth estimation, our system enables drones to infer obstacle distances and make intelligent navigation decisions in real-time. This aligns with recent work in end-to-end deep learning for unknown environment exploration \cite{9721080} and vision-based frameworks for autonomous navigation \cite{mademlis2024vision}. We scaled the depth map values to real-world measurements through a combination of experimental and programmatic methods using camera parameters. 

We introduce a novel dataset featuring an indoor representation of a city within a $5\times4$ meter area. This controlled environment allows us to train and test autonomous flight capabilities, including obstacle avoidance, scene surveying, and safe landing procedures. We used this environment to generate segmentation masks using classical machine learning algorithms and these masks were then used as training data for a neural network for semantic segmentation \cite{marcu2019towards, girisha2019semantic}. Additionally, for the second level of our system, we used the flight data to train an unsupervised network for estimating relative depth from RGB images, building upon recent advancements in monocular depth estimation \cite{bian2019unsupervised, godard2019digging}. 

To validate our approach, we conducted extensive real-world tests that demonstrated the effectiveness of incorporating depth prediction in terms of both navigation distance and execution speed. To further analyze and refine our algorithms, we created a digital-twin environment, a virtual replica of the real-world environment, where we repeated experiments and tested our methods. This approach was necessary to ensure our algorithm behaves as expected over longer sequences of experiments, since real-world testing is more expensive and carries the inherent risk of accidents. These simulations clearly confirmed the practical benefits of our methodology and highlighted the advantages of combining semantic segmentation with depth prediction.

Recognizing the need to reduce dependency on external systems (e.g., laptops connected via Wi-Fi), we developed an optimized version of our system. By distilling all mathematical algorithms, control decisions, monocular depth estimation, and semantic segmentation into a single neural network, we achieved a self-supervised system capable of running on embedded devices, including small drones. Our tests confirmed the feasibility of this approach, demonstrating that a relatively lightweight network (1.6 M parameters) could replicate the functionality of the more complex system with comparable performance of 87.5\% mission success rate while enabling fully autonomous operation without external computational dependencies.	

Our study integrates existing vision-based concepts, resulting in a rigorously tested product for robustness, reliability, and safety. This builds upon recent advancements in integrating perception, guidance, and navigation using deep learning for autonomous drone operation \cite{8299437, 9543598}. By focusing on monocular vision-based control, we contribute to developing more accessible and versatile autonomous drone systems.

Furthermore, our research aligns with the growing interest in unsupervised and self-supervised learning methods for depth estimation and semantic segmentation \cite{bian2021auto, sun2022sc}. These approaches potentially increase drone adaptability to new environments by not requiring extensive ground-truth data \cite{huang2017structure, 8877728}.

The main contributions of this work are summarized as follows:
\begin{itemize}
\item \textbf{Vision-Only Navigation:} A multitask system combining semantic segmentation and monocular relative depth prediction together with an adaptive scale factor for autonomous drone navigation.

\item \textbf{3D Virtual Safety Corridor with Metric Depth:} A real-time obstacle avoidance system using five-plane safety geometry that enables dynamic flight path optimization.

\item \textbf{Controlled Indoor Dataset:} A comprehensive dataset featuring $39648$ frames across multiple scene configurations in a custom $5\times4$ m indoor environment dataset for training and evaluating vision-based drone tasks in reproducible conditions.

\item \textbf{Real-World and Simulated Validation:} Extensive tests in both real and digital-twin environments to ensure reliability and robustness, demonstrating strong correlation between the real and simulated environment across 130 total flight tests.

\item \textbf{End-to-End Control Learning for Embedded Implementation:} A lightweight (1.6M parameters) self-supervised network for onboard execution on small drones that learns complete flight policies from demonstrated data, achieving autonomous operations and 87.5\% mission success rate. The training is self-supervised learning to improve adaptability and reduce reliance on labeled data.
\end{itemize}
    
This research demonstrates practical solutions for vision-only autonomous drone navigation in controlled environment, with implications for applications in warehouses, indoor surveillance, theme parks and with a few tweaks, such as state-of-the-art semantic segmentation, even in outdoor environments. By addressing fundamental challenges in metric depth estimation and computational efficiency, we advance the feasibility of deploying autonomous drones in real-world scenarios where traditional expensive sensing modalities are prohibitive.

\section{Related Work}
\label{sec:related-work}
Vision-based control for autonomous drones has evolved from classical techniques to deep learning approaches, addressing navigation, obstacle avoidance, and scene understanding across diverse applications including search and rescue \cite{10522748}, wind turbine inspection \cite{drones8040154}, and payload transportation \cite{10243043}. These systems offer significant benefits in safety, efficiency, and cost-effectiveness \cite{9364669, 10.1145/3344276}.

A primary challenge in autonomous flying robots is achieving accurate depth perception with minimal sensor input. While multi-sensor approaches combining depth sensors with IMUs and GPS provide reliable spatial awareness \cite{pirvu2021depth, licuaret2022ufo}, they increase system cost and complexity. Monocular approaches \cite{bian2019unsupervised, bian2021unsupervised} reduce hardware requirements but lack metric scale. Our system addresses this by converting non-metric depth predictions into real-world metric depth using semantic segmentation and geometric cues, enabling precise obstacle avoidance without additional sensors.

Neural network-based approaches have shown promise in diverse environments. Forest navigation systems estimate drone orientation and distance using vision-based networks \cite{smolyanskiy2017toward}, while others optimize speed and latency by outputting collision probability and steering commands \cite{navardi2022optimization}. However, many methods depend on extensive supervised training or global state estimation. Our approach leverages self-supervised learning to eliminate labeled depth data requirements while enabling real-time deployment on small drones.

Semantic segmentation enables comprehensive environmental understanding for safer operations. Recent advances include deep learning models for gate detection in drone racing \cite{8299437}, vision-based frameworks for autonomous cinematography \cite{mademlis2024vision}, and lightweight algorithms achieving autonomous flight through gate sequences at speeds up to 1.5m/s \cite{LI2020103621}. While previous systems demonstrate segmentation for navigation \cite{kakaletsis2021computer, a17040139}, they often treat segmentation and depth estimation separately. Our method integrates these tasks in a multitask learning framework, validating the approach through both real-world tests and digital-twin simulation.

Deep reinforcement learning has achieved impressive results in drone racing, reaching speeds up to 8m/s \cite{foehn2022alphapilot} and 60km/h with physical quadrotors \cite{9636053}. Recent work focuses on learning deep sensorimotor policies using contrastive learning \cite{10341805}, while exploration in unknown environments uses end-to-end techniques with mixed reality frameworks \cite{9721080}. Multi-task regression-based learning approaches have been developed for UAV flight control in unstructured outdoor environments \cite{8771214}, and monocular SLAM methods enable real-time trajectory estimation with semi-dense reconstruction \cite{8098709}. Unlike these planning-based systems requiring global optimization, our approach performs reactive navigation based on real-time visual understanding, simplifying deployment in constrained environments.

Classical image processing techniques remain relevant, with RGB filtering for object tracking \cite{bi2013implementation} and optical flow-based background subtraction combined with Mask R-CNN \cite{rs16050756}. State estimation methods including Extended Kalman Filtering \cite{liu2020real, karpenko2015visual, bachrach2009autonomous, eller2019advanced} and Monte Carlo Localization \cite{eller2019advanced} face limitations in nonlinear scenarios, prompting research into Nonlinear Model Predictive Control (NMPC) and Distributed Model Predictive Control (DMPC) for swarm coordination \cite{9562281}. Our work adopts a fully deep learning-based multitask architecture, bypassing explicit state estimation through real-time scene understanding for improved scalability on resource-constrained platforms.

Self-supervised CNN approaches for indoor navigation predict distance-to-collision for safe movement \cite{kouris2018learning}, while lightweight perception modules enable efficient operation on nano-drones \cite{10532299, 8715489}. Supervised approaches have associated images with collision probability in traffic scenarios \cite{loquercio2018dronet}. Energy efficiency considerations have led to optimized approaches like E2EdgeAI \cite{9996805} and unified frameworks for edge deployment \cite{mohan2023unified}. Our embedded solution distills complete perception and decision pipelines into self-supervised networks, enabling onboard execution without external computing infrastructure while achieving energy-aware design goals.

Vision Transformers have shown superior performance in high-speed obstacle avoidance \cite{bhattacharya2024vision, dosovitskiy2020image}, while spatio-temporal architectures enhance drone-to-drone detection \cite{10161433}. Dynamic obstacle tracking with trajectory prediction represents significant advances in robust navigation \cite{Zhong_2024_WACV}. Advanced applications include trajectory planning frameworks with SE(3) planners \cite{9543598}, game theoretic planners for two-player drone racing \cite{9112709}, automated tour management for long-haul flights \cite{9779119}, swarm coordination without inter-agent communication \cite{9363551, 8798720}, and Voronoi-based neighbor selection for maintaining cohesive behavior \cite{9732989}. Our system achieves comparable navigational capabilities using compact CNN architectures with metric reasoning via depth and semantic cues, offering practical advantages in real-time embedded scenarios.

Recent research emphasizes unsupervised approaches for environmental classification \cite{huang2017structure} and domain adaptation between simulated and real environments \cite{8877728}. In monocular depth estimation, methods like \cite{bian2019unsupervised, bian2021unsupervised, bian2021auto, sun2022sc} demonstrate growing interest in self-supervised approaches, while others have utilized GPS data with structure-from-motion algorithms \cite{pirvu2021depth, licuaret2022ufo} for metric estimation. Binocular stereo approaches generate depth from single images \cite{godard2017unsupervised, godard2019digging}. For semantic segmentation, datasets like RuralScapes \cite{marcu2019towards} and specialized UAV applications \cite{girisha2019semantic, 10490368, nigam2018ensemble} provide foundations for training. Our work reduces reliance on extensively labeled data by using self-supervised learning for the depth model directly from flight data and a knowledge distillation approach for the segmentation model, which learns from a weakly-supervised teacher that generates frames from flight data.

These advancements collectively contribute to vision-based autonomous drone control, addressing challenges in navigation, safety, environmental interaction, and computational efficiency across various applications. Our approach uniquely combines semantic understanding with metric depth estimation in a unified, lightweight framework suitable for resource-constrained platforms.

\section{Methods}
\label{sec:methods}
Our methods are based on efficient and innovative algorithms that combine the tasks of semantic segmentation with monocular metric depth estimation. The algorithm is executed on a laptop computer that communicates with the drone via WiFi. The best performing method then serves as a teacher for a tiny student network (1.6M parameters) that learns the mission end-to-end in an unsupervised manner. This distilled network can be deployed directly onboard small drones with limited computational power, demonstrating its embedded capabilities. The methods for semantic segmentation and metric depth estimation are also fast, cost-effective, and efficient while maintaining good performance, enabling real-time execution on a standard laptop.

\subsection{Architecture and Setup}
\label{subsec:methods-architecture}
Our autonomous drone system is designed to operate in diverse environments, capable of flying independently to specified locations and performing surveillance-like tasks. The drone can autonomously generate and follow optimal paths, leveraging semantic segmentation to identify safe landing zones in critical situations. Furthermore, by using depth estimation for each pixel together with the semantic segmentation, our system enables obstacle avoidance and precise timing of control actions.

\begin{figure*}[!t]
    \centering
    \includegraphics[width=\linewidth]{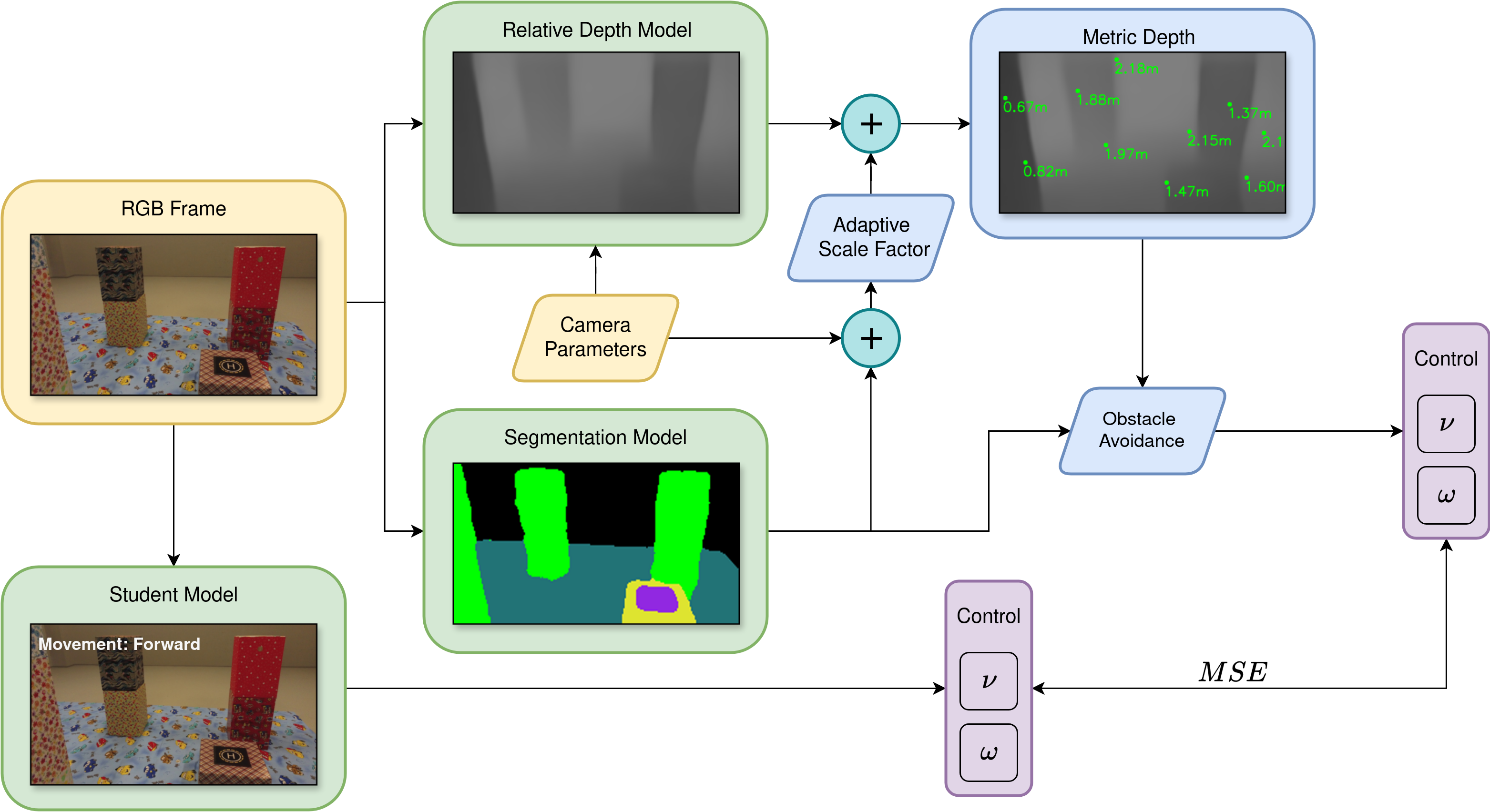}
    \caption{System architecture for our vision-based autonomous control. The framework processes RGB frames through multiple pathways: a relative depth model and segmentation model provide geometric and semantic understanding, while camera parameters enable an adaptive scale factor computation for obtaining metric depth. These components are combined to generate drone control commands (translations $\nu$ and rotations  $\omega$) which are then used to train a compact student neural network through knowledge distillation which employs a mean squared error (MSE) loss for training.}
    \label{fig:drone-scheme}
\end{figure*}

The architecture integrates real-time processing of visual data for intelligent decision-making while maintaining operational flexibility and safety through manual override capabilities. The modular design allows for the integration of advanced algorithms for semantic understanding and depth perception, crucial for autonomous navigation in complex environments.

To implement these ideas we created an indoor environment inside our laboratory room from the Precis building at the University "Politehnica" of Bucharest. This guaranteed us a controlled environment that is 5 meters tall, and $10\times10$ meters floor area. The room has featureless walls, seen in Figure \ref{fig:dataset-imgs}, and floor and those initially posed challenges for the drone stability. The lack of visual features impaired the optical flow calculations used internally by the downward-facing camera for position holding, causing the UAV to drift towards soft shadows. To mitigate this issue and delineate the operational boundaries, we introduced a textured carpet ($5\times4$ meters). This addition provided abundant visual features, significantly enhancing the drone's stability during fixed-point hovering and movement. This controlled environment, measuring $5\times4$ meters, serves as a testing scene for autonomous flight capabilities, including obstacle avoidance, scene surveying, and safe landing procedures. To validate our methods and be able to extend our testing we replicated this laboratory room together with its features in a simulated environment. This allowed us to have more confidence in our real-world experiments.

\subsection{Dataset}
\label{subsec:methods-dataset}
To evaluate the autonomous capabilities of our UAV in accessible environments, we developed a custom dataset featuring an indoor representation of an urban landscape. The dataset was collected in the laboratory room setup presented in Section \ref{subsec:methods-architecture}. 

To create a more complex testing environment, we incorporated eight cardboard box obstacles which were wrapped with distinct patterns and one designated landing point. These boxes were wrapped in patterned paper to provide the necessary visual features for recognition and collision avoidance. The tops of the obstacles were adorned with distinct patterns to facilitate roof recognition, crucial for safe navigation and landing procedures. The main idea of these cardboard box obstacles is to simulate a city landscape. 

Data collection involved flying the drone around the scene in multiple sessions and configurations, each lasting about $2-3$ minutes. Flights consistently began from the helipad box, with the camera angled at $-25$ to $-30$ degrees to capture a forward-looking view. We ensured balanced coverage of each obstacle to provide sufficient training data for our neural networks. Landing sequences were captured by positioning the drone above the helipad and adjusting the gimbal to 90 degrees, providing a top-down perspective.

The dataset consists of four different configurations of the scene, designed for the network to learn to estimate different types of segmentations and depth in different relations with other obstacles. These four configurations are composed of $2-3$ videos per scene forming a total of nine videos. The nine videos have a total of $39648$ frames that are about $22$ minutes of flight. The videos are divided into seven videos for training and two for validation. For real-world testing we used two different scene configurations from which we trained to evaluate the generalization of our methods. 

To further test the drone capabilities, we replicated the real-world environment, complete with its obstacles, using Unreal Engine and the Parrot Sphinx simulator \cite{WhatisPa31:online}. This simulation served as a digital-twin of the laboratory. First, we created $3D$ meshes in Blender \cite{blender2024} and applied textures using real-world images. The simulation can be launched through either a custom-built Unreal Engine application or by populating an existing one with the corresponding meshes at specified coordinates, thus being able to replicate different scenes configurations. 

We replicated four scenes from the real-world dataset to match the coordinates for data collection. Additionally, two testing scenarios were recreated as levels within the simulation, enabling us to test our algorithms and methods efficiently. This approach boosted our confidence in transitioning to real-drone testing while minimizing the risk of potential crashes.

\subsection{First Level - Semantic Segmentation}
\label{subsec:methods-segmentation}
The semantic segmentation is adapted to the fact that we know the textures of important objects, but in the outdoor real-world this module can be swapped to a state-of-the-art network such as \cite{jain2023oneformer, cheng2022masked, bolya2019yolact}. In our case some textures are known, a case possible in diverse controlled environments such as warehouses, halls, or theme parks for children. This allows for a simpler system that requires minimal human intervention and training data to create a SVM teacher model. This model then generates masks used to train a small student neural network for segmentation. In a more complex environment a state-of-the-art network could be used as a teacher for the smaller network, in the literature this was proven to work and to be efficient, distilling a bigger neural network which is more general to a smaller more focused on a type of real scenes \cite{marcu2023self}.

The semantic network is vital to understand the scene at a semantic level in order to find the base to return and land safely. It is also an initial solution to avoid obstacles without the need of estimating the depth. Furthermore, semantic understanding is also crucial for estimating metric depth which can be used for a better obstacle avoidance. Thus, semantic segmentation helps with both finding the returning to base point, or any target in fact, but also to estimate the metric depth from a relative depth map, this being the first task that needs to be completed.

\begin{figure}
    \centering
    \includegraphics[width=\linewidth]{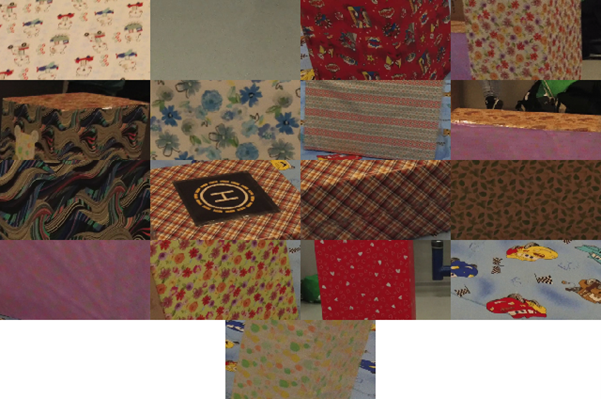}
    \caption{Representation of all present class textures in the scene. On the first row, the second image represents the background, which is represented by walls and floors. On the third row, the second and third images represent the helipad box, the zone in which the drone initiates and concludes its mission. On the fourth row, the final image represents the carpet flooring which is our fly zone.}
    \label{fig:dataset-semantic-classes}
\end{figure}

The carpet class is essential as it represents the safe zone for our drone to fly and is later used with relative depth estimation to obtain metric depth measurements. Other important classes include the background and boxes, which the drone must avoid to prevent collisions. The helipad class is separated from the other boxes since it represents the designated safe landing spot. Initially, the helipad serves as the primary point of interest for the drone, as this is where the drone must return safely after completing its mission. The drone primary objectives are to fly around the scene while avoiding collisions, remain within the scene boundaries, and return safely to the base (i.e., the helipad). For mission planning, we divided the semantic classes into four categories: the helipad, the boxes (obstacles), the carpet (safe to fly zone) and the background (everything outside the carpet boundaries). 

After selecting the classes to be considered for a complete understanding of the environment, we proceeded to select regions of interest from each class across 4 different scenes, which will be further used for the depth estimation task. From each scene, approximately $7-9$ images were chosen from each of the $17$ classes. Each class of interest was captured from different angles, distances, and lighting conditions across the images in each scene. To optimize processing, we resized all images to $30\%$ of their original size using area interpolation.
 
\begin{figure*}[!t]
    \centering
    \includegraphics[width=\textwidth]{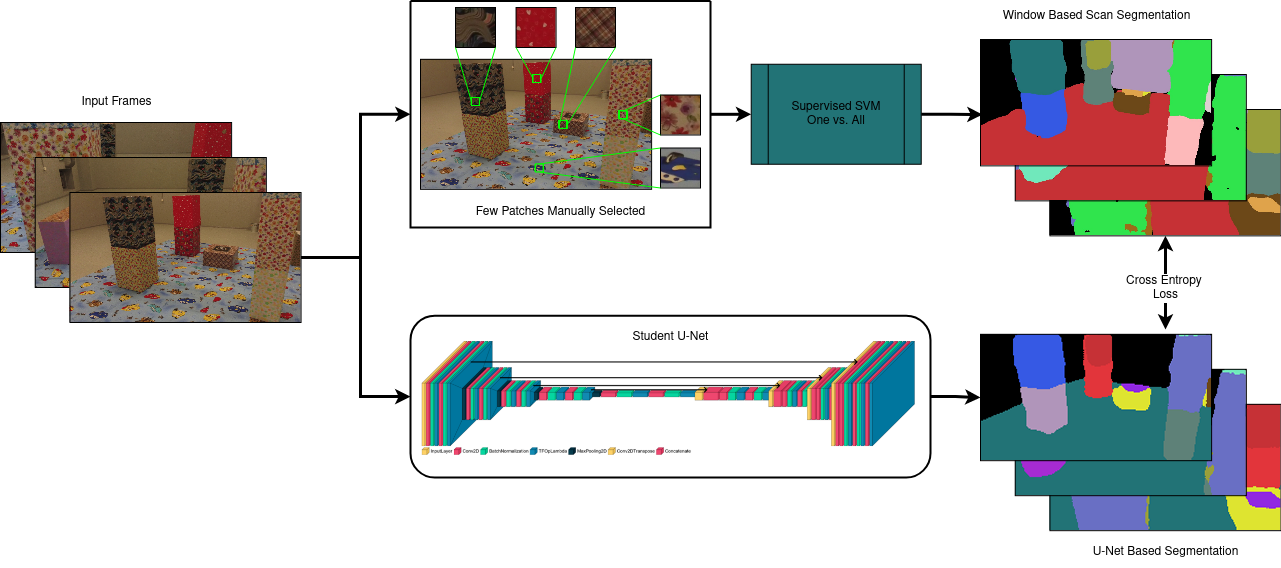}
    \caption{The architecture of the knowledge distillation process for the segmentation model. From the RGB frames a few coordinates for each class were manually selected and processed into $40\times40$ pixel HSV patches. These patches train an SVM model with One vs. All classification to generate class specific masks for each frame. Due to the whole SVM computational limitations for real-time processing, a U-Net is subsequently trained using the SVM-generated masks as ground truth teacher data, enabling faster segmentation for real-time applications.}
    \label{fig:teacher-svm-student-unet}
\end{figure*}

A total of $6484$ points of interest were obtained from the four scenes, and tests were done with patches from only one scene, then from two scenes and so on to see the improvement of the model in an iterative way. The points of interest represented the coordinates on X and Y of the Cartesian coordinates of the image, and the points were transformed into patches of $40\times40$ pixels in order to gather more features from a class. Each each of those patches was then transformed into Hue Saturation Value (HSV) patches with $90$ bins for hue, $64$ bins for saturation, $36$ bins for value, which would then be concatenated into a single array. The HSV features would be used to train a Support Vector Machine (SVM) model that uses One VS Rest classifier to classify patches into their respective classes. 

Using this method, the inference for a single frame took approximately $6-7$ seconds with the teacher patch-based SVM color histogram approach. However, to achieve real-time performance, we implemented a knowledge distillation approach. We used the SVM model to generate segmentation masks for every 5th frame in our dataset. These masks, along with the original RGB frames, were then used to train a U-Net \cite{ronneberger2015unetconvolutionalnetworksbiomedical} architecture as a faster student model, as shown in Figure \ref{fig:teacher-svm-student-unet}. 

The teacher network created a dataset consisting of $3474$ frames scaled to a resolution of $224\times112$ pixels, using area interpolation to better preserve image information. With $17$ masks per frame (corresponding to $17$ classes), this yielded $59058$ masks as training input. The classes were selected based on important objects in the scene, including the background, carpet, two classes for the helipad and its surrounding box, and one class for each other unique box or roof structure. The architecture implemented for real-time image segmentation is based on the U-Net architecture.

Using semantically segmented imagery from the built model, we detect the designated return point (helipad) by identifying its two components, the "H" sign and the yellow-red striped box pattern surrounding it. Through component analysis with calibrated thresholds, we evaluate potential component pairs based on spatial proximity, connectivity, and geometric compactness metrics to filter out false positives. Once the most promising pair is identified, we can assume that the helipad has been detected in the frame, this approach also operates in real-time.

\subsection{Second Level - Depth Estimation}
\label{subsec:methods-depth}
The depth estimation task is essential for our obstacle avoidance system. While pretrained networks that can detect monocular depth, they provide relative depth estimation compared to other objects in the scene, rather than a metric depth map. Metric depth estimation enables safer control, by allowing us to estimate the time to impact based on the known speed of a drone and the distance to obstacles. Without access to exact metric depth, movement becomes less safe. Our tests confirmed that the relative depth provided by monocular networks cannot be reliably converted to metric depth using a single constant scale factor learned from the scene. 

The monocular depth estimation was trained on our custom laboratory dataset, with various models tested. We primarily focused on unsupervised SC-Depth variants found in \cite{bian2019unsupervised}, \cite{bian2021unsupervised}, \cite{bian2021auto}, and \cite{sun2022sc}. We chose these models due to the lack of ground-truth depth data in our initial dataset and relatively small in number of parameters (e.g., $14M$ to $34M$). These depth estimation algorithms use camera intrinsic parameters to compensate for lens distortion effects and improve depth measurement accuracy. Furthermore, knowledge of camera intrinsic parameters aids in camera calibration, which is essential for precise depth estimation. The configuration with the lowest reprojection error was chosen as input for the depth estimation neural network and subsequent scale factor computation.

Using our dataset, we fine-tuned a SC-depth variant model on the frames of each selected scene to infer the depth maps of the images. This process provides values between $0$ and $1$ for each pixel in the image. With these values, we can determine the mean, median, and minimum values from a region of interest. 

We propose two methods for computing the scale factor. The first method involves dividing the ground truth depth values in our region of interest by the predicted relative depth values generated by the network. The second method uses an equation with two unknowns, one being the scale factor and the other a shift parameter. The latter equation can be expressed as:

\begin{equation}
    \label{eq:1}
    ScaleFactor \cdot Predicted_{depth} + Shift = GroundTruth
\end{equation}

Where the ground truth was measured from the drone camera to marked points on the box. Then, for each frame where we have the ground truth, we generate one equation following the form above, resulting in N equations for N frames. To solve this system of equations, we apply the method of least squares to determine the optimal values for the two unknown variables: the scale factor, and the shift.

For each method used to acquire the predicted depth maps in the region of interest, we compute the mean, median, and minimum values. For each of these values, we calculate the corresponding scale factor, and in the case of the second method, we also determine the shift parameters. These scale factors (and shifts for the second method) can then be used to establish a suitable scale factor for metric prediction in each scene.

\subsection{Enhanced Depth - Adaptive Scale Factor}
\label{subsec:methods-adaptive-scale-factor}
Initial attempts to convert relative depth into metric depth values yielded unsatisfactory results, with significant errors in our controlled environment. This led us to develop an adaptive scale factor that adjusts to the scene in real-time. Our solution leverages camera intrinsic and extrinsic parameters and a segmented carpet that serves as a known plane parallel to the drone. This allows us to compute the distance from the camera to random points on the carpet, thus enabling the calculation of an adaptive scale factor in real-time with minimal computational overhead.

To implement this solution, we require the frame with the segmented floor mask, the rotation matrix $R$ with respect to the $y_c$ axis, the translation vector $T$, and the camera parameters $K$ that were initially computed for the relative depth estimation network.

We can compute the rotation matrix $R$, where we know the steering angle of the gimbal $\theta$ as: 
\begin{equation}
    \label{eq:2}
    R = \begin{bmatrix}
        \cos(\theta) & 0 & \sin(\theta) \\
        0 & 1 & 0 \\
        -\sin(\theta) & 0 & \cos(\theta)
    \end{bmatrix}
\end{equation}

The translation vector is defined as below, since we know that the UAV is positioned at $1.5$ meters above the ground plane:
\begin{equation}
    \label{eq:3}
    T = \begin{bmatrix}
        x_t \\
        0 \\
        0
    \end{bmatrix}
\end{equation}

Then, for a point with pixel coordinates $x$ and $y$, we create its homogeneous representation $P$:
\begin{equation}
    \label{eq:4}
    P = \begin{bmatrix}
        x \\
        y \\
        1
    \end{bmatrix}
\end{equation}

With these parameters, we can compute the camera coordinates using the inverse of the intrinsic parameters:
\begin{equation}
    \label{eq:5}
    P_c = K^{-1} \cdot P
\end{equation}

Using the above calculations, we can compute the point coordinates with respect to the world coordinate system:
\begin{equation}
    \label{eq:6}
    P' = R \cdot P_c + T
\end{equation}

Next, we define the vector $v$ with the notation of:
\begin{equation}
    \label{eq:7}
    v = P' - T \implies v = \begin{bmatrix}
        x' \\
        y' \\
        z'
    \end{bmatrix} - \begin{bmatrix}
        x_t \\
        0 \\
        0
    \end{bmatrix}
\end{equation}

The coordinates of the corresponding point on the ground in the 3D world coordinates can be computed as: 
\begin{equation}
    \label{eq:8}
    P'' = T + \lambda \cdot v \implies \begin{bmatrix}
        0 \\
        y'' \\
        z''
    \end{bmatrix} = \begin{bmatrix}
        x_t \\
        0 \\
        0
    \end{bmatrix} + \lambda \cdot \left (\begin{bmatrix}
        x' \\
        y' \\
        z'
    \end{bmatrix} - \begin{bmatrix}
        x_t \\
        0 \\
        0
    \end{bmatrix}\right)
\end{equation}

Where, the value of $\lambda$ represents a scaling term that can be expressed as:

\begin{equation}
    \label{eq:9}
    x_t + \lambda \cdot (x' - x_t) = 0 \Rightarrow \lambda = \frac{-x_t}{x' - x_t}
\end{equation}

Using these equations, we can estimate the 3D world coordinates of ground points. The distance between the camera and a ground point is calculated as the Euclidean distance between $P''$ (the ground point) and $P'$ (the camera position). By repeating this process for multiple points in an image, we obtain a set of distances that allows us to compute a scale factor that adapts to the environment and drone parameters such as translation and rotation.

We computed the scale factor for each scene using two different approaches. In the first method, we computed the median value from all frames withing that specific scene. For the second method, we used least squares regression across all test scenes.

\begin{figure}
    \centering
    \includegraphics[width=\linewidth]{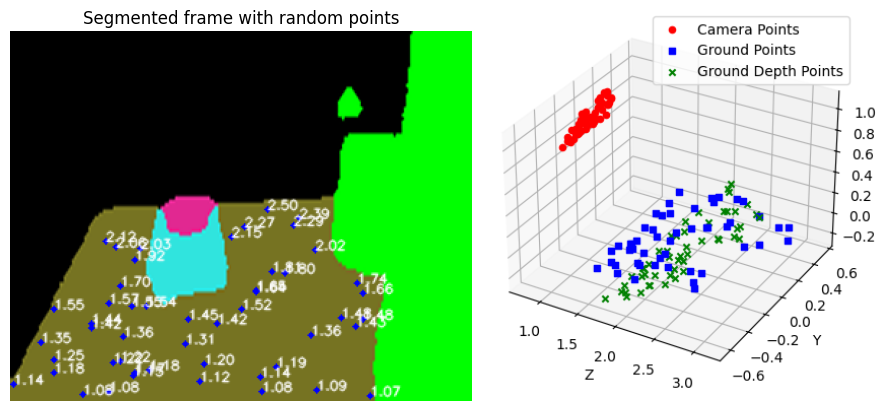}
    \caption{On the left side is the 2D image with $50$ randomly selected points from the ground segmented plane that are computed as Euclidean distances between the ground point $P''$ and the camera position $P'$. On the right side is the 3D representation of the points.}
    \label{fig:scale-factor-3d-points}
\end{figure}

\begin{figure}
    \centering
    \includegraphics[width=\linewidth]{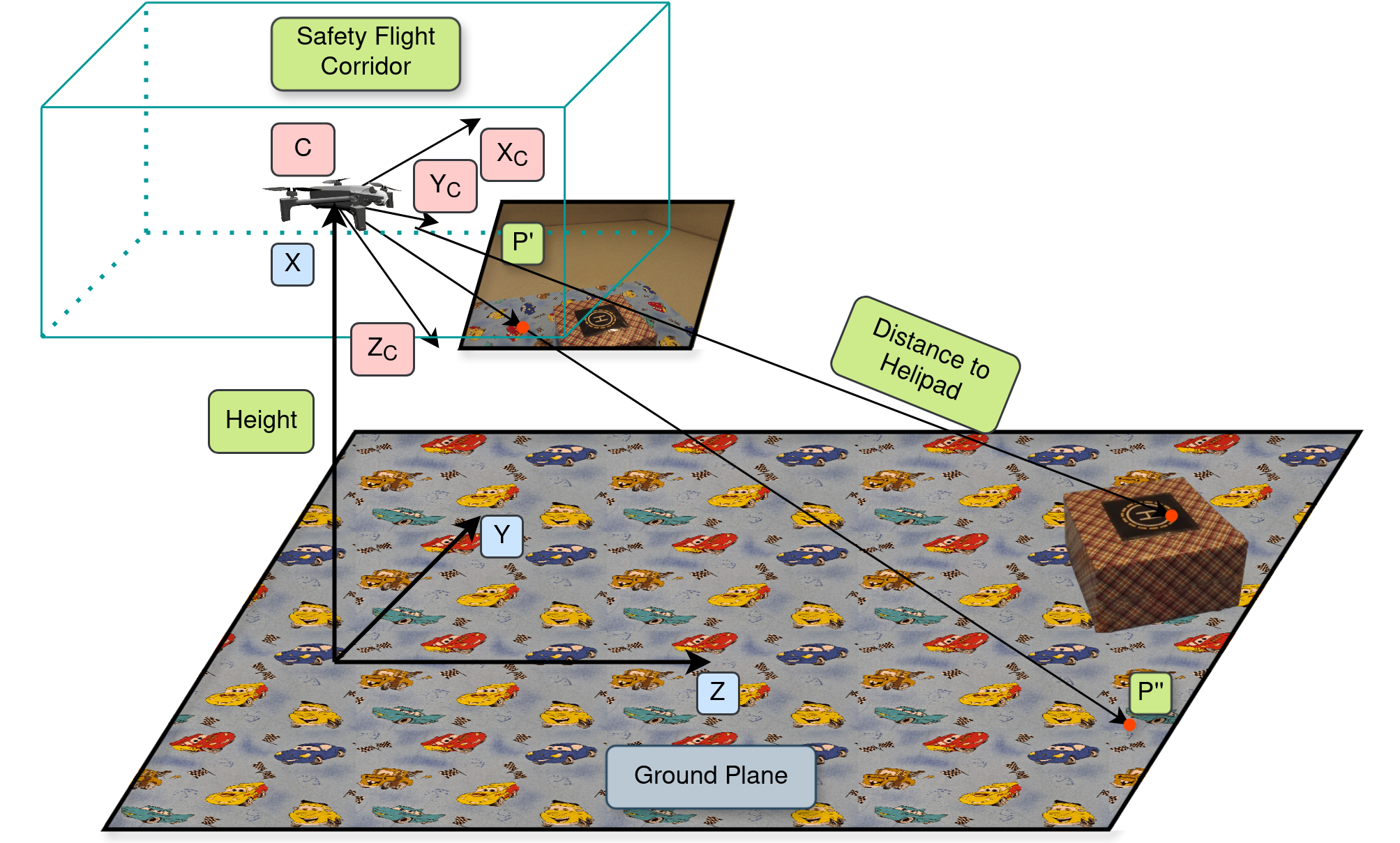}
    \caption{3D representation of the drone camera system within a virtual safety corridor. The diagram illustrates the geometric relationship between the drone coordinate system $C$ with axes $X_c, Y_c, Z_c$ and the world coordinate system $X, Y, Z$, showing how image points $P$ are projected on the ground plane $P''$. The distances (e.g., 0.5m  for left/right planes, 2m for the frontal plane). For any point in the camera field of view, the system determines which safety plane would be intersected first based on minimum Euclidean distance calculations using the scaling parameters defined in Equations \ref{eq:10A} - \ref{eq:10E}. The segmented carpet pattern on the ground plane serves as a reference surface for real-time scale factor computation and metric depth estimation.}
    \label{fig:scale-factor-box}
\end{figure}

This method can be used to compute a scale factor at each frame and compute a metric distance from a depth map whenever necessary. However, for safety reasons, we propose a 3D virtual safety box, as shown in Figure \ref{fig:scale-factor-box}, that aligns with the drone's safety thresholds. This box comprises five planes at different distances relative to the drone camera. For example, the left and right planes may be set at $0.5$ meters from the camera, while the frontal plane is at $2$ meters.

Following the previous formulation for computing the 3D point coordinates, we can derive a similar set of formulas:

\begin{subequations}
    \label{eq:10}
    \begin{align}
        \lambda_{low} = \frac{x_{low} - x_t}{x' - x_t} \label{eq:10A}\\
        \lambda_{high} = \frac{x_{high} - x_t}{x' - x_t} \label{eq:10B}\\
        \lambda_{left} = \frac{y_{left} - y_t}{y' - y_t} \label{eq:10C}\\
        \lambda_{right} = \frac{y_{right} - y_t}{y' - y_t} \label{eq:10D}\\
        \lambda_{front} = \frac{z_{front} - z_t}{z' - z_t} \label{eq:10E}
    \end{align}
\end{subequations}

To create the virtual box, we constrain the scaling value $\lambda$ to be greater than 1. We then compute $P''$ for each plane and calculate the Euclidean distances. The minimum distance determines which plane the point in the image will intersect first in 3D space.

To further refine the calculations, we introduce two additional parameters for more realistic results. The first parameter is a shift along the $Z$ axis in the translation vector, since the camera position in 3D world coordinates is slightly farther from the center of the drone. This results in the new translation matrix $C$:

\begin{equation}
    \label{eq:11}
    C = \begin{bmatrix}
        x_c \\
        0 \\
        z_c
    \end{bmatrix}
\end{equation}

The second parameter is the focal length $f$, which is considered when computing the real 3D projection of the camera in world coordinates, resulting in $P\prime_{center}$. 
\begin{equation}
    \label{eq:12}
    P'_{center} = C + (P' - C) \cdot f
\end{equation}

\subsection{Autonomous Control}
\label{subsec:methods-autonomous-control}
We implemented two autonomous control methods that both handle obstalce avoidance, scene exploration, and return to base. One method relies just on segmentation data, while the other combines segmentation with metric depth data.

The main mission of the drone is to fly autonomously around the scene, avoiding obstacles and returning to the landing point safely and efficiently in both real-world and digital-twin environments. These algorithms can be transferred to outdoor environments with minimal modifications.

To test the algorithms, we proposed a control mechanism where the drone flies in a stochastic manner while avoiding obstacles and remaining within the scene boundaries. After a predetermined duration (e.g., $90$ seconds in our case), it commences the return to home procedure, during which the drone continues to fly randomly until it detects the helipad. Since the environment is tiny, this step requires minimal time and no additional heuristics. Upon detecting the helipad, the drone initiates the landing procedure.

Initially, to implement and test the landing procedure, the drone was positioned at a random point within visual range of the helipad to ensure that it can be detected. The drone would take-off and adjust its heading to steer towards the helipad using a PID controller to minimize rotational error and ensure accurate orientation. Once the helipad is centered in the frame, the drone moves towards it. When sufficiently close (i.e., when the helipad was detected previously and is not detected for $15$ consecutive frames), the camera is then oriented perpendicular to the ground plane (i.e., \(\pi / 2\) radians from its default orientation). The drone then translates upward along the Z-axis to improve detection accuracy. 

\begin{figure}
    \centering
    \includegraphics[width=\linewidth]{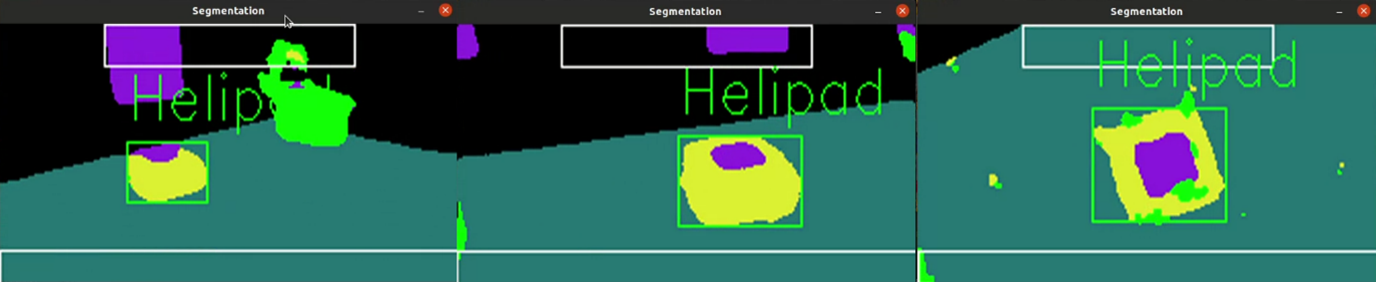}
    \caption{Sequential frames from a drone mission demonstrating autonomous helipad detection, centering, and landing. The semantic segmentation algorithm identifies the helipad (shown in purple/magenta) and its surrounding box (yellow), while the drone autonomously adjusts its position and orientation to center the helipad within the camera field of view. The progression shows the helipad moving from off-center positions in the left frames to a centered position, with the drone adjusting its gimbal orientation to achieve an orthogonal view of the landing surface.}
    \label{fig:control-find-helipad}
\end{figure}

With the helipad visible in the camera frame, the drone aligns the center of the camera frame with the center of the detected helipad. It then moves $15$ centimeters forward, corresponding to the distance between the drone center and the camera, before landing. This sequence is illustrated in Figure \ref{fig:control-find-helipad}.

\subsubsection{Self-flying with segmentation only}
\label{subsubsec:self-flying-seg-only}
Initially, a method based exclusively on semantic segmentation was implemented to test the capabilities of our drone. This method was designed to be lightweight and fast. Our proposed algorithm enables the drone to fly independently and safely in its environment. After analyzing the flight data, we determined that safe flight can be achieved by considering the total number of points within two rectangles: one at the top of the image representing the obstacles based on the gimbal tilt, and one at the bottom to verify that the drone remains within the scene. 

The movement criteria are as follows:
\begin{itemize}
    \item If the lower region has more than $1600$ points and the upper region has fewer than $900$ points, the drone moves forward.
    \item If the lower region has more than $1600$ points and the upper region has more than $900$ points, this indicates an obstacle ahead. The algorithm divides the upper region in half and rotates toward the side with fewer points.
    \item If the lower region has fewer than $1600$ points and the upper region has fewer than $900$ points, this means the drone is about to exit the scene. The algorithm divides the upper region in half and rotates toward the side with more carpet points.
    \item If none of these conditions are met, the drone performs a U-Turn ($\pi$ radians rotation).
\end{itemize}

We analyze the number of pixels in specific classes within the upper and lower rectangles. This approach improves obstacle avoidance efficiency and enables safe landings by recognizing objects in the video stream, identifying helipads, and initiating flight sequences without collisions. In emergencies, the drone can land safely on a carpet-class surface.

The autonomous flying algorithm avoids obstacles and prevents exiting the scene (e.g., flying outside the carpet area). It uses semantic segmentation to check for specific classes within regions of interest. For the upper part of the scene, we count pixels from any box class, excluding helipad and its box. The lower part of the scene requires counting the number of pixels from carpet, obstacle box, and helipad classes. 

When an action is needed (i.e., when thresholds are met), the regions of interest are divided into two equal parts. The drone steers toward the side with more target class pixels and continues moving forward. The upper scene threshold is set at $1000$ pixels, while the lower scene threshold is $1600$. 

\subsubsection{Self-flying with segmentation and metric depth}
\label{subsubsec:self-flying-seg-metric-depth}
Because the segmentation-only approach relies on small, sequential movements and waits for each movement to complete before processing the next frame, this approach can result in slower operations. However, by implementing an adaptive scale factor alongside a depth estimation algorithm, we enhanced the autonomous control capabilities. With knowledge of the metric distance from the drone to obstacles and a defined safety corridor, we can improve the speed at which the drone executes movements. The system predicts how far it can move in a single action without collision. For instance, instead of moving forward $30$ centimeters for each adjustment, the drone can now advance further when it is safe to do so, optimizing its path and reducing the number of sequential commands required. 

The algorithm operates on the same primary principles as the previous method \ref{subsubsec:self-flying-seg-only}: maintaining scene boundaries and performing adaptive obstacle avoidance.

The first principle involves using knowledge of the segmented carpet to keep the drone within the flight area, similar to the segmentation-only method. By identifying the segmented carpet, the autonomous system can continuously verify that it remains within the designated boundaries. If the number of pixels corresponding to the carpet class falls below a predefined threshold, the system initiates a rotation maneuver to reorient itself back toward the scene.

The second principle implements a 3D virtual safety box comprising five planes positioned at predetermined distances from the drone camera. This box is not a physical object but rather a conceptual space created by five distinct planes positioned at specific metric distances from the drone. The lateral planes are positioned at $0.5$ meters, while the frontal plane extends $2$ meters ahead, with additional upper and lower planes completing the safety corridor. For any given point in the depth map, the algorithm computes intersection points with all safety planes using Equations \ref{eq:10A}-\ref{eq:10E}, determining collision risks based on the minimum Euclidean distances. This minimum distance value is crucial. If the minimum distance to an obstacle is less than the predtermined safety corridor boundaries, the autonomous collision avoidance system is triggered. This direct, metric based feedback allows the drone to make real-time decisions, such as rotating to a different side if an obstacle is detected within a plane, enabling the algorithm to assess collision risks in 3D space rather than relying on 2D pixel analysis. For example, if the obstacle projected point has a minimum Euclidean distance to the right lateral plane, we know an obstacle is on the right side of the drone, thus it will rotate left. This ensures the drone steers away from the immediate threat while also favoring a path with a clear line of sight.

The integration of metric depth fundamentally transforms the movement decision logic. Rather than executing fixed 30-centimeter increments, the algorithm calculates maximum safe travel distances for each navigation step. When the frontal safety plane indicates clear space beyond $2$ meters, the drone advances up to $1$ meter in a single movement. Additionally, the system incorporates camera offset corrections through the translation matrix $C$ and focal length parameter $f$ as defined in Equations \ref{eq:11} and \ref{eq:12}, accounting for the physical displacement between the camera sensor and the drone center to improve the spatial accuracy of the safety computations.

This approach requires knowledge of the segmented carpet to convert relative depth into metric depth, as discussed in Section \ref{subsec:methods-adaptive-scale-factor}. The carpet class remains necessary for scene boundary maintenance, utilizing the same approach from Section \ref{subsubsec:self-flying-seg-only}, but enhanced with metric distance information for improved precision and decision-making capabilities. To return to the helipad, we also used the previous method of flying randomly until it is detected.

\begin{figure*}
    \centering
    \includegraphics[width=\textwidth]{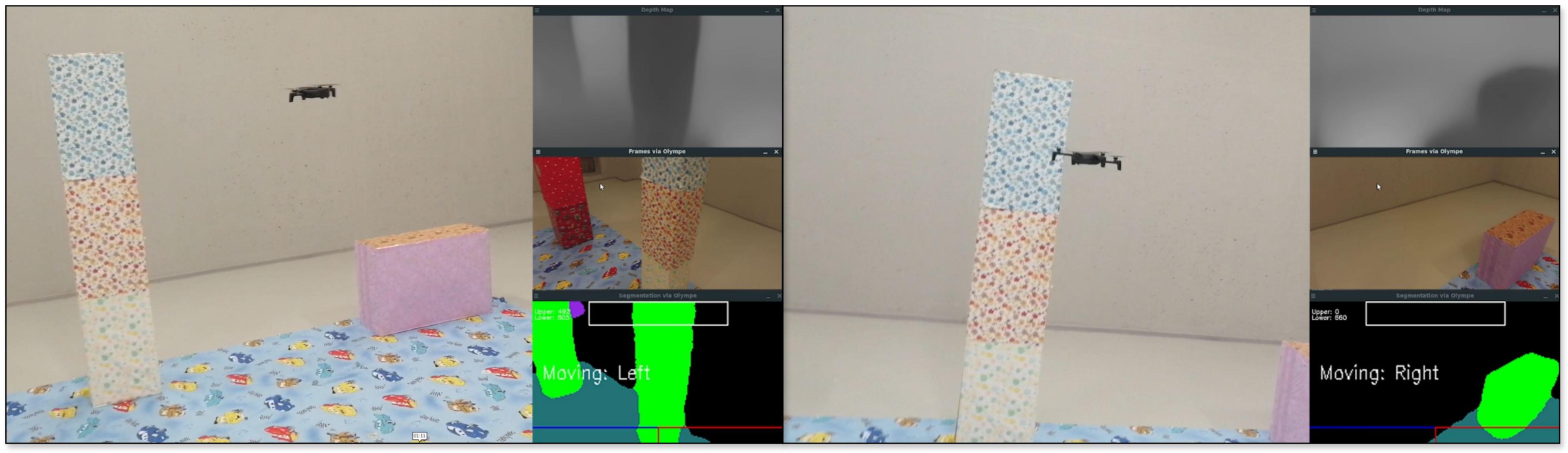}
    \caption{Real-time autonomous flight demonstration showing a multi-task approach for obstacle avoidance and scene boundary management. Each panel displays four synchronized views: the raw drone camera feed (middle left), an external observer perspective showing the drone in the environment (right), semantic segmentation output (bottom left), and depth estimation visualization (top left). The left panel side is the representation of avoiding the obstacle which is the tall box. The right panel side is the representation of the steering to prevent scene boundary leaving.}
    \label{fig:control-avoid-obstacles-stay-in-scene}
\end{figure*}

\begin{figure}
    \centering
    \includegraphics[width=\linewidth]{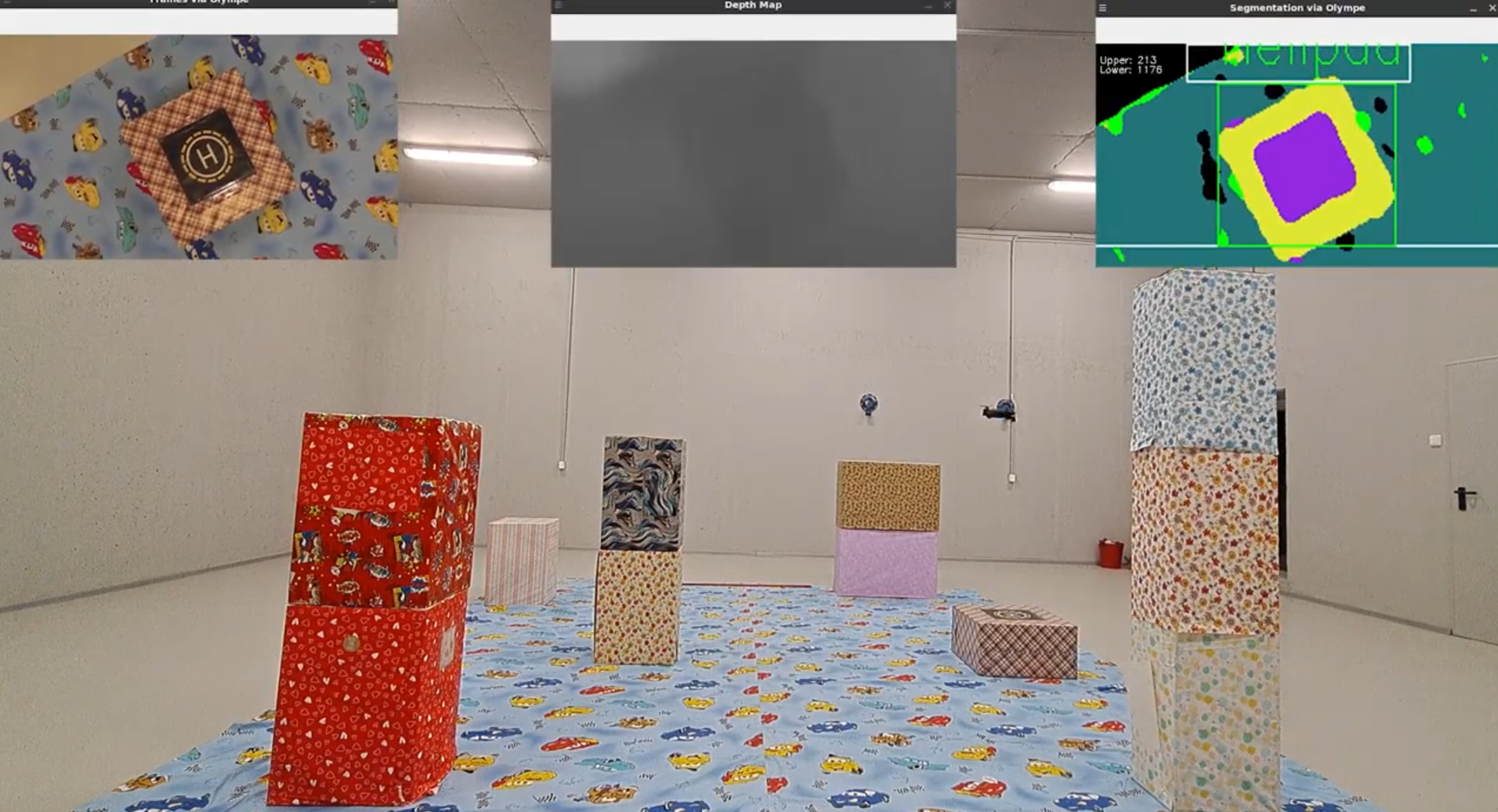}
    \caption{Example of landing with all the processed scenes and the point of view from a third perspective.}
    \label{fig:control-find-helipad-with-actor}
\end{figure}

\subsection{Student Control Network}
\label{subsec:methods-student-control-network}
The previous approaches require two medium-sized neural networks (i.e., 25M parameters for segmentation and 14M parameters for depth estimation) and some logic for movement prediction. While this configuration is sufficient for running on a low-cost laptop, it becomes computationally demanding when implemented on edge devices such as small drones. To address this challenge, we implemented a compact tiny neural network with only $1.6$M parameters that learns from flight data collected during our experimental runs from Methods \ref{subsubsec:self-flying-seg-only} and \ref{subsubsec:self-flying-seg-metric-depth}.

For data acquisition, we employed the methods used in previous approaches to fly around the scene for approximately 90 seconds before initiating the helipad-finding state. The collected data included metrics necessary to test our methods (as discussed in Section \ref{sec:experiments}), frames, along with their corresponding movement decisions and timestamps.

The network takes as input the current image and, if available, the previous two frames from within a one second temporal window distance. The output is the predicted movement, which can be evaluated against our previously established methods. The architecture resembles U-Net but omits the upward convolutional layers, instead directly predicting movement through dense layers.

\section{Experiments and Results}
\label{sec:experiments}
All experiments were conducted using a Parrot ANAFI drone equipped with a $4K$ resolution camera, controlled via the Olympe SDK framework. This SDK provides access to drone movement controls and video stream data, facilitating seamless transition between testing and deployment phases. Ground processing was performed on a laptop with an Intel Core i7-8750H processor, 16GB RAM and NVIDIA GTX 1050Ti GPU, communicating with the drone via WiFi. The laboratory environment maintained consistent lighting conditions without natural light interference. The digital-twin was created in Unreal Engine 4 with Parrot Sphinx framework \cite{WhatisPa31:online} and meshes for the population of the environment were made in Blender \cite{blender2024}.

We performed camera calibration using the checkerboard method \cite{zhang2002flexible}. To estimate camera parameters, we captured $350$ images of a checkerboard with $14 \times 25$ squares, each measuring $20mm$. Since the depth model uses $320 \times 256$ frames and the drone camera captures at $4K$ resolution, we downscaled the images to $320 \times 256$ using area interpolation to preserve information. We exhaustively searched for detectable corner pattern configurations, identifying four viable options within the checkerboard dimensions. The corners represent the number of squares for width and height that can be detected, while the number of images represents the actual count of images where squares matching the corner configuration can be found. The reprojection error indicates how accurately the camera intrinsics would perform when applied to reprojection, the lower the error, the better the parameters found. For the digital-twin drone, the camera parameters are already available. All experiments used consistent camera settings: 30 FPS recording and gimbal angle fixed at $-25^{\circ}$ to $-30^{\circ}$ during flight phases.

\subsection{First Level - Semantic Segmentation - Evaluation and Setup}
\label{subsec:experiments-segmentation}
To evaluate the classifier for a single frame, we parsed the image using a grid-based approach with a step size of 5 pixels and extracted $40\times40$ pixel patches at each grid point. We chose this sampling strategy instead of parsing every pixel to reduce computational intensity. Inference was then performed on each patch, generating a binary mask where the target class appeared as white pixels agains a black background. This grid based sampling approach resulted in a dot-like appearance of the classified objects due to gaps between sampled locations. To address this, we applied morphological dilation to fill the missing areas and restore object continuity. While this process introduced some additional pixels at object margins, the trade-off was acceptable given the significant computational savings.

However, since segmentation relies on HSV, occasional misclassifications may occur where a class is incorrectly identified as another. In order to avoid such errors in our frames, we employed a connected component analysis to retain only the largest component of each class segment. This strategy was particularly effective for classes that occupied a smaller portion of the frame (i.e., smaller objects). For background and carpet this step was skipped since there can be multiple instances of these classes without them being connected.

\begin{figure}
    \centering
    \includegraphics[width=\linewidth]{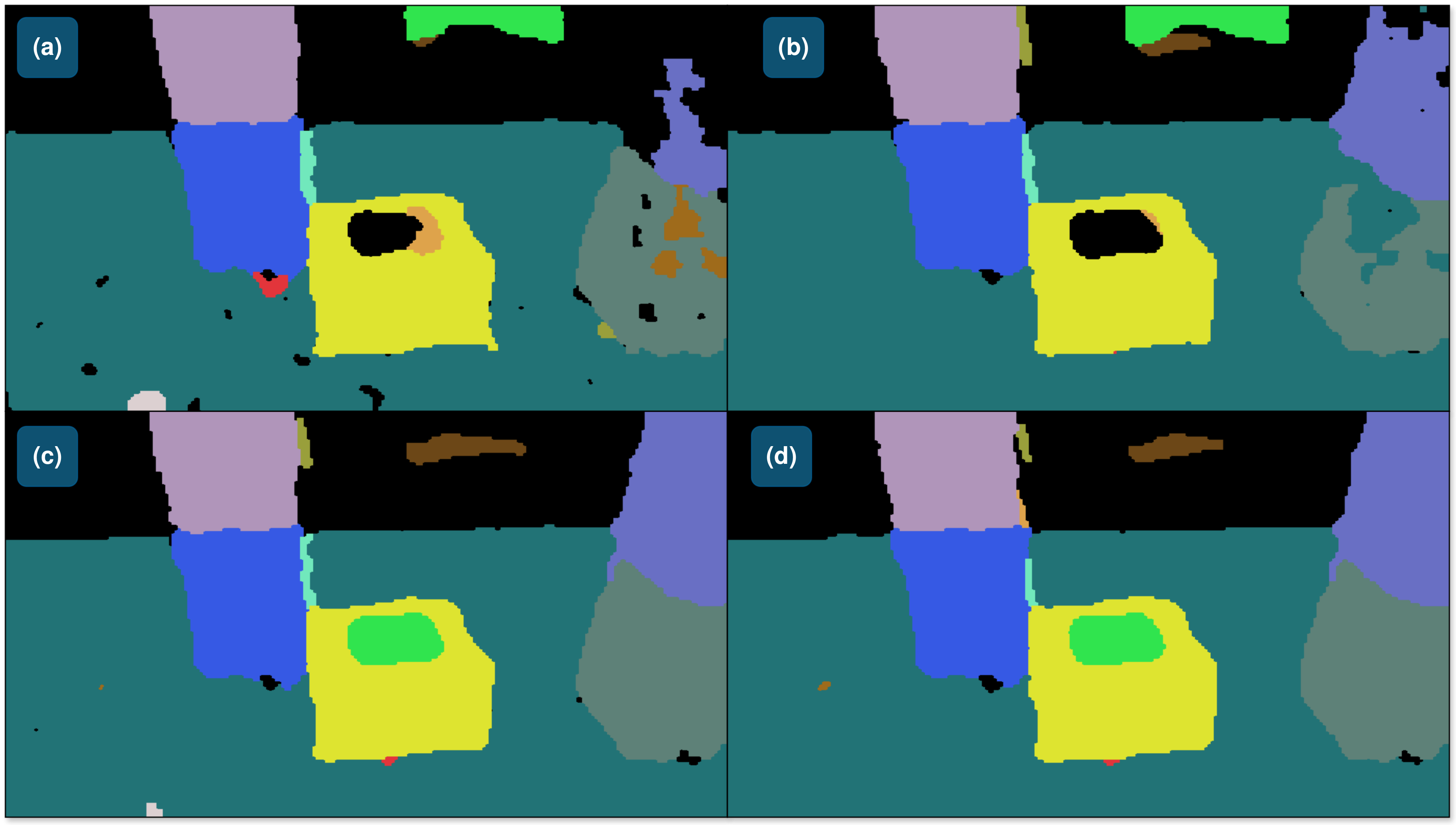}
    \caption{Visualization of SVM color-based segmentation performance with incremental training data. (a) Model trained with patches from one scene, (b) trained with patches from scenes 1-2, (c) trained with patches from scenes 1-3, and (d) trained with patches from all four scenes, demonstrating improved segmentation quality with increased scene diversity.}
    \label{fig:semantic-segmentation-results-svm}
\end{figure}

\begin{figure}
    \centering
    \includegraphics[width=\linewidth]{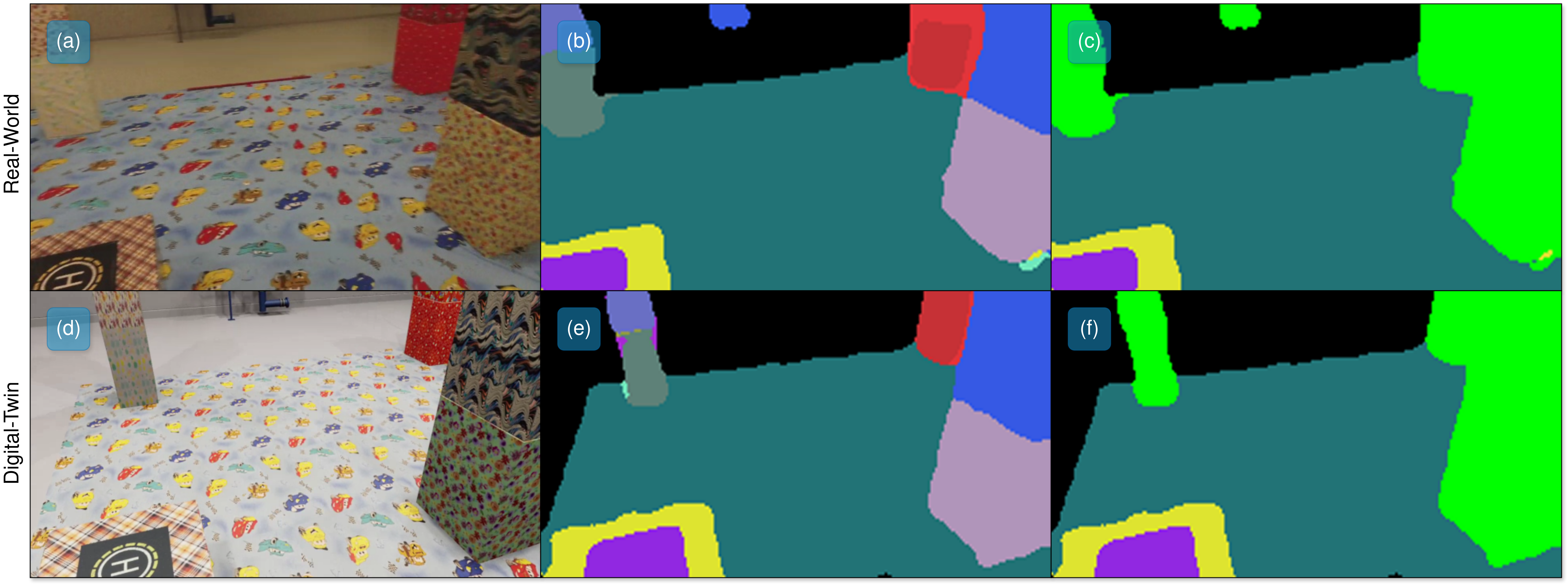}
    \caption{Student U-Net segmentation results on an unseen scene for flight testing validation. Top row represents images from real-world and bottom images from digital-twin environments with input images (a) and (d). The network outputs are showing in (b, e) for individual class segmentation and (c, f) simplified obstacle classification where all obstacle classes are combined into a single class for navigation purposes.}
    \label{fig:semantic-segmentation-nn-all-sep-vs-all-boxes-combined}
\end{figure}

Despite these optimizations, inferring a frame still took approximately seven seconds even with a multi-threaded solution. Given our need for real-time processing during flights we considered implementing a neural network based on a U-Net architecture. This approach would leverage the masks generated by our current method to perform real-time class segmentation. Essentially, this can be seen as a knowledge transfer algorithm where the U-Net neural network acts as the student and the One-vs-All Support Vector Machine serves as the teacher. Visual examples with student-teacher inference can be seen in Figure \ref{fig:semantic-segmentation-nn-all-sep-vs-all-boxes-combined} for both real-world and simulated environments. This approach not only was faster but also yielded more accurate and consistent segmentation results.

\begin{figure*}
    \centering
    \includegraphics[width=\textwidth]{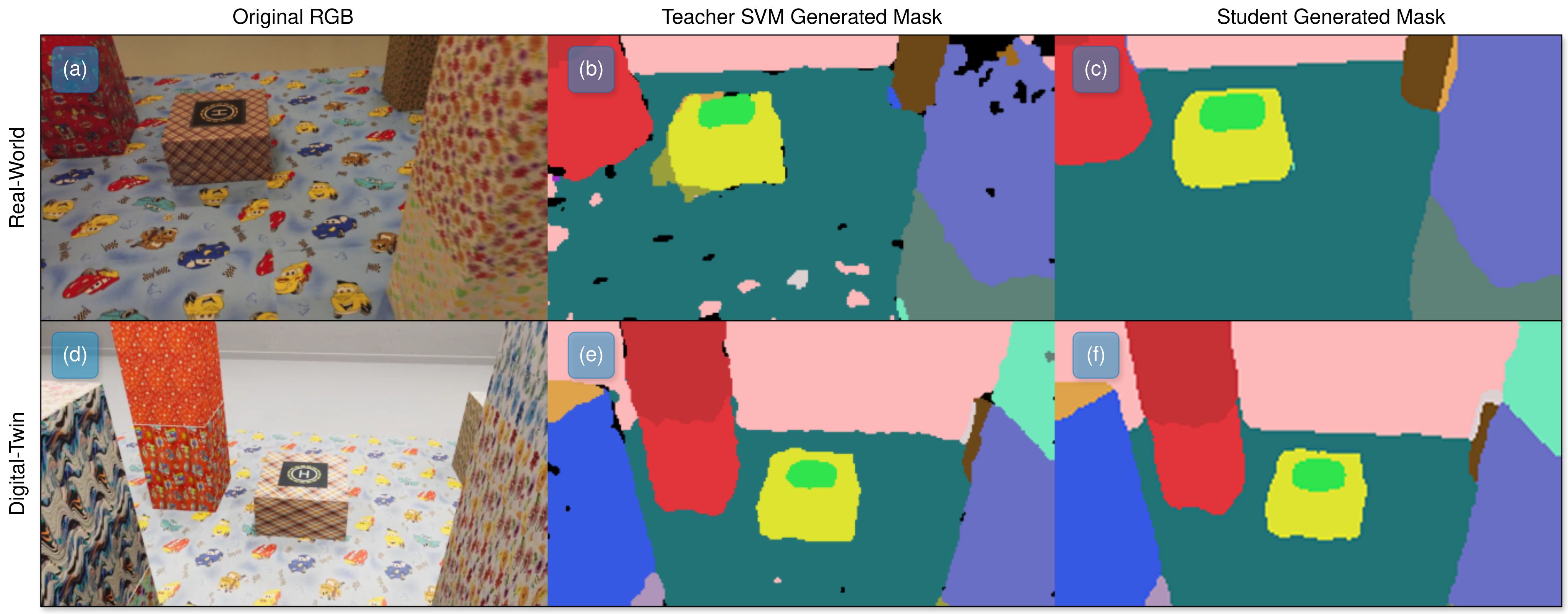}
    \caption{Comparison of segmentation results on real-world (top row) and simulated (bottom row) environments. (a,d) Original RGB images, (b,e) teacher SVM-generated masks using patch classification, and (c,f) student U-Net-generated masks trained on SVM outputs, demonstrating the knowledge distillation effectiveness across different environments.}
    \label{fig:sim-vs-real_mask_vs_nn}
\end{figure*}

Several augmentations were applied to the dataset, in a similar manner to RuralScapes \cite{marcu2019towards}, meaning that Hue value was rotated randomly by up to 5 degrees, the Saturation was adjusted by a range of 2\% and the Value was adjusted by a range of 6\%. In order to help the model learn features better random translations and rotations were applied for the images. Stronger HSV augmentations, than the one mentioned, proved to harm the ability of the neural network to learn to semantically segment the image, the chosen values mentioned yielding better results.

In the case of losses and outputs for the network, several combinations were tried. The worst performing was the combination least squares error with a 3 color output head, matching a color-based segmentation, as it failed to learn a reasonable representation of the segmentations, which resulted in an unusable model. The second combination tried was a replication of RuralScapes Binary Cross Entropy with Dice loss over a 17-mask segmentation. Dice loss computes the overlap between two segmentations, and it is generally considered better behaved on unbalanced segmentation datasets. Finally, the method which provided the best results with the highest accuracy, and which was closer to the original masks was the Categorical Cross Entropy over the 17-mask segmentation.

\subsection{Second Level - Depth Estimation - Evaluation and Setup}
\label{subsec:experiments-depth}
To evaluate our monocular depth estimation algorithm, we used two testing methods. The first involved using video sequences from scenes not included in the training data. These videos were designed to check whether the scaled depth estimates remained consistent across different scenarios. The second method consisted in running the algorithm in real-time during a drone test flight on both simulator and real-world.

After training the model, we evaluated its performance by inferring depth maps for the training videos. This process yielded both depth maps and their corresponding visualizations. We also tested the model on scenes it had not encountered before (e.g., the same laboratory environment but with boxes arranged differently). The results were promising, producing outputs that appeared reasonable given the features present in the frames.

\begin{figure}
    \centering
    \includegraphics[width=\linewidth]{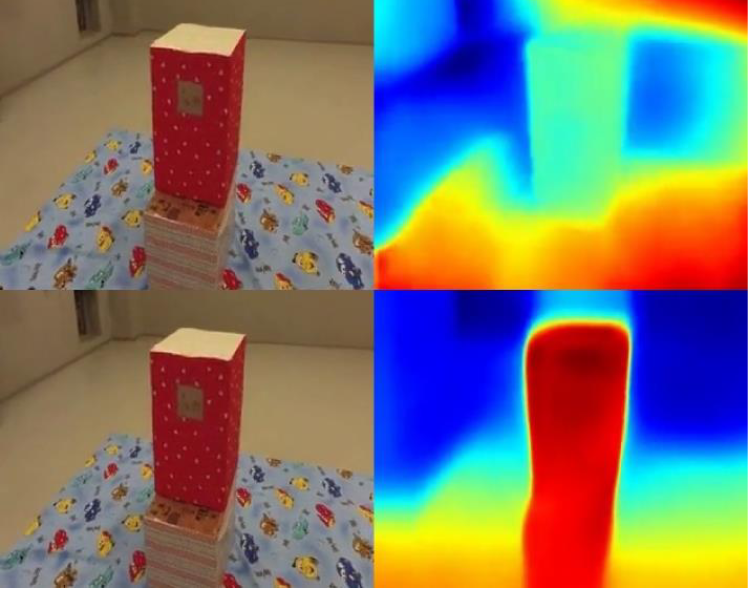}
    \caption{Comparison between the model trained on the Nyu Depth V2 dataset (top) and the model trained on our laboratory dataset (bottom). Some inconsistencies can be observed in the NYU Depth V2 model, as the box in reality is close to the camera.}
    \label{fig:depth-nyu-vs-ours}
\end{figure}

Considering that Nyu Depth V2 \cite{silberman2012indoor} is a dataset for depth estimation and semantic segmentation based on indoor scenes, we tested the model trained on this dataset.  As expected, the depth estimates were inconsistent when applied to our laboratory environment, as the model had never been exposed to such features during training. In some cases, the predicted depths were highly inaccurate, objects appeared very far away (blue color in the visualization \ref{fig:depth-nyu-vs-ours}) despite being close in reality. While the model occasionally produced decent depth predictions for certain features, the overall inconsistency led us to conclude that this dataset was unsuitable for our specific application.

\begin{figure}
    \centering
    \includegraphics[width=\linewidth]{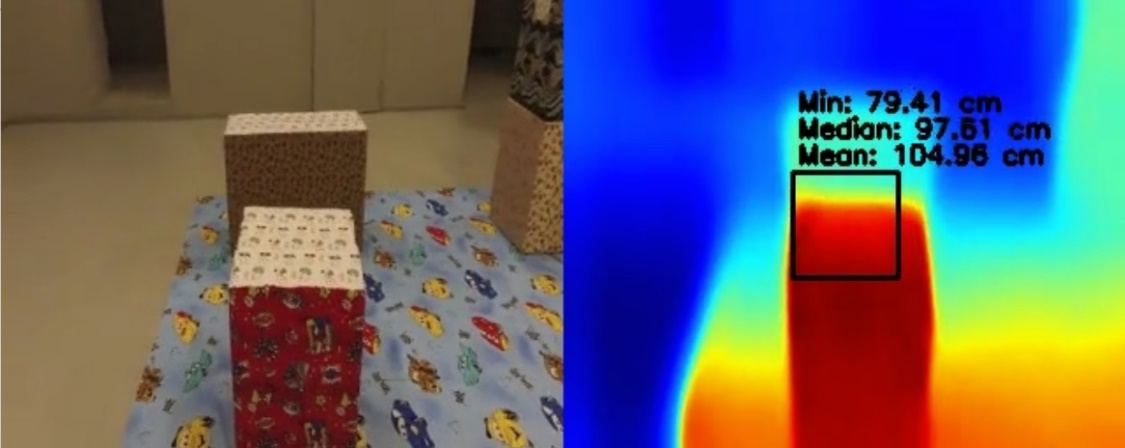}
    \caption{RGB image on the left and the depth map image visualization on the right. Min value represents the lowest value of a pixel in the region of interest (i.e., the black square), and the median and mean represents the median and the mean respectively of each pixel in ROI.}
    \label{fig:depth-scaled}
\end{figure}

We concluded that the model performs well when the scale factor is applied to closer measurements, specifically, within 50 centimeters of ground truth measurements. However, it achieves only about a 30-centimeter error in detecting depth metrics when tested on scenes with features farther away and no relative closer features. In such cases, the errors reach approximately 100 centimeters, which is unacceptable. This behavior led us to consider an adaptive scale factor. To address this issue, we propose using the camera intrinsic parameters and a segmented carpet (a ground plane) to compute the distance from the camera to random points on the carpet. By doing so, we can calculate an adaptive scale factor in real-time. This approach results in a lightweight and fast metric depth scaling solution.

\subsection{Adaptive Scale Factor}
\label{subsec:experiments-scale-factor}
To evaluate the adaptive scale factor implementation, we designed a controlled experiment using intrinsic and extrinsic camera parameters. The intrinsic parameters were mentioned in the depth training section \ref{subsec:methods-depth}. For the extrinsic parameters, we needed the rotation from Equation \ref{eq:2} of the camera with respect to the ground plane and the translation from Equation \ref{eq:3} and \ref{eq:11} with respect to the same plane. Ground truth distances were measured using a roulette meter measure, taking measurements from the drone camera position to specific points on segmented plane regions and obstacles points of interest. Each location was measured 10 times, and the median value was used to minimize human measurement error.

The choice of 50 random sampling points was determined through preliminary testing, where we evaluated sampling densities from 10 to 100 points. Results showed that 50 points provided an optimal balance between computational efficiency and measurement stability. Points were sampled using uniform random distribution across the segmented carpet region, excluding boundary pixels within 10 pixels of mask edges to avoid artifacts. For each test frame we computed the median scale factor rather than mean to reduce sensitivity to outliers caused by segmentation errors or depth estimation noise. Using these scale factors, we then calculated the metric depth for regions of interest where ground-truth distances were available, enabling us to compute an error metric. Additionally, we explored estimating the scale factor using least squares regression. However, this method yielded slightly worse results compared to our adaptive approach.

\begin{table*}[!ht]
\centering
\caption{Performance comparison of different patch sizes and models.}
\label{tab:adaptive-vs-sota}
\begin{tabular}{l l c c c c c}
\toprule
\textbf{Patches} & \textbf{Model} &
\textbf{Mean Error [m]} $\downarrow$ & \textbf{Median Error [m]} $\downarrow$ &
\textbf{RMSE [m]} $\downarrow$ & \textbf{STD [m]} $\downarrow$ & \textbf{Correlation Coef} $\uparrow$ \\
\midrule
No Patch   & MoGe      & 0.2102 & 0.0642 & 0.5088 & 0.4637 & 0.2330 \\
           & Depth Pro & 0.2111 & 0.0781 & 0.4798 & 0.4637 & 0.1816 \\
           & Ours      & 0.0740 & 0.0597 & 0.0988 & 0.0655 & 0.8342 \\
\midrule
3x3 Median & MoGe      & 0.2095 & 0.0651 & 0.5059 & 0.4608 & 0.2362 \\
           & Depth Pro & 0.2123 & 0.0781 & 0.4836 & 0.4348 & 0.1821 \\
           & Ours      & 0.0736 & 0.0589 & 0.0980 & 0.0647 & 0.8353 \\
\midrule
5x5 Median & MoGe      & 0.2140 & 0.0646 & 0.5117 & 0.4652 & 0.2431 \\
           & Depth Pro & 0.2188 & 0.0789 & 0.4981 & 0.4478 & 0.1846 \\
           & Ours      & 0.0744 & 0.0578 & 0.0994 & 0.0659 & 0.8308 \\
\midrule
7x7 Median & MoGe      & 0.2184 & 0.0644 & 0.5220 & 0.4745 & 0.2379 \\
           & Depth Pro & 0.2234 & 0.0800 & 0.5021 & 0.4500 & 0.1826 \\
           & Ours      & 0.0751 & 0.0623 & 0.1000 & 0.0661 & 0.8299 \\
\midrule
9x9 Median & MoGe      & 0.2210 & 0.0653 & 0.5207 & 0.4718 & 0.2486 \\
           & Depth Pro & 0.2323 & 0.0805 & 0.5168 & 0.4620 & 0.1796 \\
           & Ours      & 0.0775 & 0.0621 & 0.1025 & 0.0671 & 0.8246 \\
\bottomrule
\end{tabular}
\end{table*}

Since we know the ground truth metric values for our scene, such as the length, width, and height of our obstacles, we can evaluate our adaptive scale factor metric depth prediction and compare it against state-of-the-art model results such as Depth Pro \cite{Bochkovskii2024:arxiv} and MoGe \cite{wang2024moge, wang2025moge2}. Table \ref{tab:adaptive-vs-sota} presents the error results based on the Euclidean distances between two points representing the dimensions of an obstacle. The pairs of points were manually selected from 100 images across all scenes, yielding a total of 661 measurements. We report the mean, median, root mean squared error (RMSE), standard deviation, and correlation coefficient for five approaches: direct point selection without patches, and patch-based methods using $3\times3$, $5\times5$, $7\times7$, and $9\times9$ neighborhoods where the median depth within each patch was used to reduce noise. The correlation coefficient measures the linear relationship between predicted and ground truth distances. It ranges from $-1$ to $1$, where values closer to 1 indicate stronger positive correlation.

\begin{figure}
    \centering
    \includegraphics[width=\linewidth]{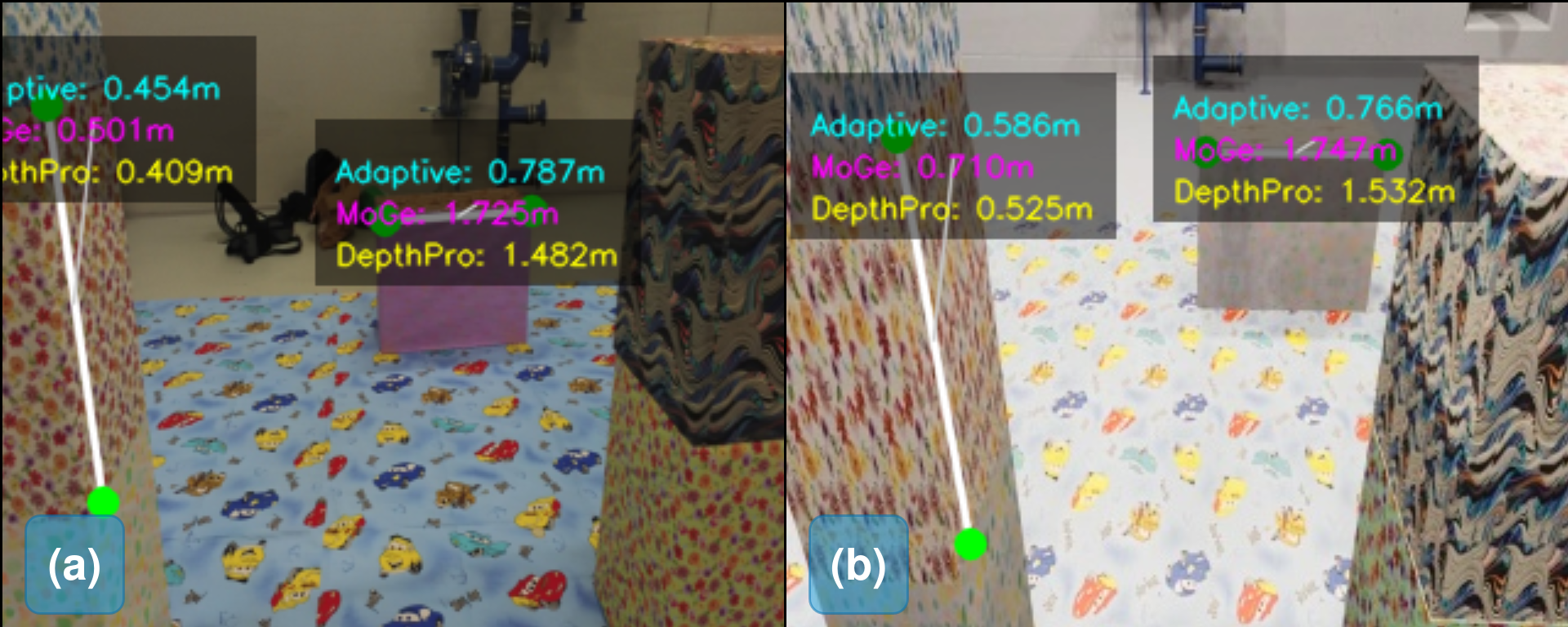}
    \caption{Comparison of our method vs MoGe and DepthPro on real-world (a) and simulated (b) environments. The distances that are closer to the camera perform similarly while the the ones further from the camera perform better for our method. Our method is highlighted in cyan color, MoGe in magenta and DepthPro in yellow.}
    \label{fig:sim-vs-depth-comparison}
\end{figure}

Notably, our model performs well on both mean and median error metrics while maintaining low standard deviation, whereas other state-of-the-art models exhibit higher mean errors and greater variability. This is especially pronounced for pairs of points that are farther apart in the image, as show in Figure~\ref{fig:sim-vs-depth-comparison}. However, our model requires knowledge of the camera parameters, the position of the ground plane, and the camera height relative to the ground plane, whereas the other methods require only the RGB image.

\begin{table}[h]
\centering
\caption{Inference time metrics for our method, MoGe and DepthPro on the dataset. The FPS is computed using the median value of the inference time.}
\label{tab:depth-inference-metrics}
\begin{tabular}{lrrrrrrr}
\toprule
\textbf{Method} & \textbf{Min} & \textbf{Mean} & \textbf{Median} & \textbf{Max} & \textbf{Std} & \textbf{FPS} \\
\midrule
Depth Pro & 2.9607 & 3.0656 & 3.0426 & 3.5957 & 0.0979 & 0.33 \\
MoGe      & 0.2873 & 0.2967 & 0.2915 & 0.3925 & 0.0155 & 3.43 \\
Ours      & 0.0193 & 0.0215 & 0.0206 & 0.7334 & 0.0206 & 48.54 \\
\bottomrule
\end{tabular}
\end{table}

Table \ref{tab:depth-inference-metrics} presents the inference time comparison across methods. Our approach achieves significantly faster performance at $48.54$ FPS compared to MoGe, $3.43$ FPS and Depth Pro $0.33$ FPS, with a median inference time of $0.0206$ seocnds per image. This computational efficiency, combined with the superior accuracy shown in Table \ref{tab:adaptive-vs-sota}, makes our method particulary suitable for real-time drone applications where both speed and precision are critical. The trade-off for this performance is the requirement for camera calibration parameters and ground plane information and the current height of the drone from the ground plane.

When testing the computed scale factor on our measured dataset with ground truth distances, the errors are shown in Table \ref{tab:table-scale-factor-errors} and are expressed in meters. This relatively large maximum error ($0.3714m$) occurred in regions with poor texture.
\begin{table}
\caption{Computed Scale Factor Errors\label{tab:table-scale-factor-errors}}
\centering
\begin{tabular}{ c c c c } 
 \toprule
 \textbf{Mean [m]} & \textbf{Median [m]} & \textbf{Min [m]} & \textbf{Max [m]} \\
 \midrule
 0.1440 & 0.1628 & 0.0003 & 0.3714 \\
 \bottomrule
\end{tabular}
\end{table}

With the adaptive scale factor established, we implemented a 3D virtual box that represents a safety flight corridor with five planes positioned at predetermined distances: lateral planes at $0.5m$, frontal plane at $2m$ and vertical planes at $\pm0.8m$ from the drone center. These distances were selected based on drone dimensions($0.24m \times 0.32m \times 0.10m$) with safety margins of $2\times$ for lateral movement and half of the carpet length.

The testing consisted of systematic evaluation of box intersection calculations across varying distances ($0.3m$ to $3m$) and different focal lengths ($0.8$ to $1.2$) and parameters withing the Equation \ref{eq:11} to assess their impact on performance. 

\subsection{Vision-Based Autonomous Flying}
\label{subsec:vision-based-autonomous-flying}
We conducted systematic evaluations of the autonomous flight capabilities using two distinct algorithms: segmentation with and without metric depth. Each trial followed a standardized approach: 90 second autonomous exploration phase followed by a return to base sequence with helipad finding and autonomous landing. 

Success criteria were defined as:
\begin{itemize}
\item Successful completion of the mission. With autonomous takeoff, obstacle avoidance during exploration, helipad detection and safe landing without any human intervention.
\item Mission failure. Any collision with obstacles, exit from designated flight area, or inability to complete the flight sequence within 300 seconds total mission timeout.
\item Crash event. Physical contact with obstacles resulting in emergency shutdown or drone damage.
\end{itemize}

Real-world experiments were performed 30 times and the simulated experiments 100 times for each method. Testing the autonomous flying system involved saving the flight data for each performed action, capturing the frames streamed from the drone, and logging the distance traveled in 90 seconds from the point of take-off and the total distance traveled until landing. Additionally, we measured the time it took for the drone to find the helipad while in the search state and the overall duration from identifying the helipad to successfully completing the mission with autonomous landing.

Flight data collection generated a comprehensive dataset used in Section \ref{subsec:student-control-network}. Real-world flights produced 24 training scenes ($14248$ frames) and 6 testing scenes ($3420$ frames), while digital-twin environment flights yielded 80 training scenes ($124542$ frames) and 20 testing scenes ($30725$ frames). Frame annotation with control commands was performed automatically during flight with manual verification for 10\% of frames to ensure data quality.

\begin{figure*}
    \centering
    \begin{minipage}{0.48\textwidth}
        \includegraphics[width=\textwidth]{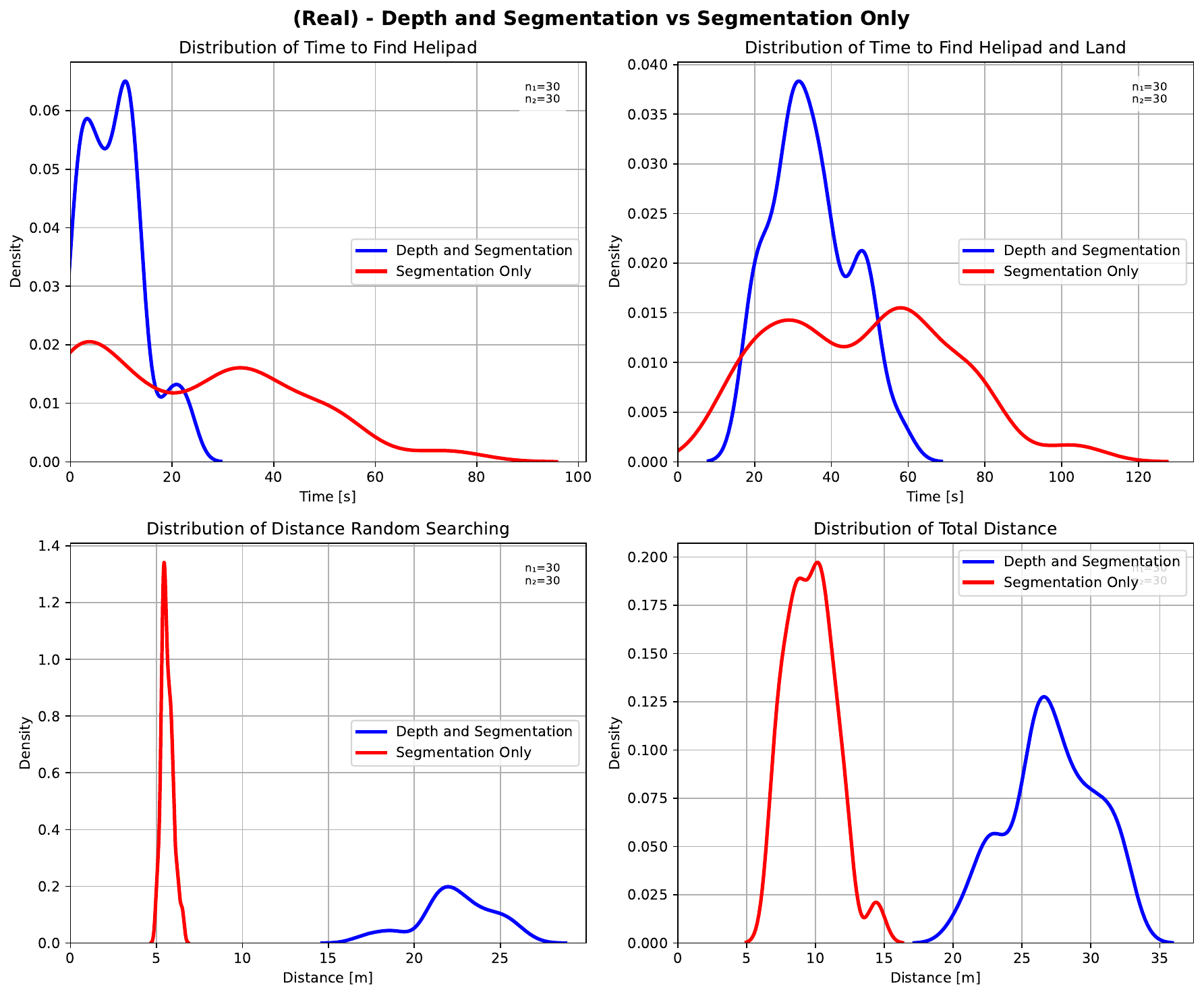} 
        \caption{Comparison of the two methods used for the real-world environment. The top row represents the distribution of time to find helipad and the bottom one, the distance traveled.}
        \label{fig:kde_comparison_depth-segmentation-tests_vs_segmentation-tests}
    \end{minipage}
    \hfill
    \begin{minipage}{0.48\textwidth}
        \includegraphics[width=\textwidth]{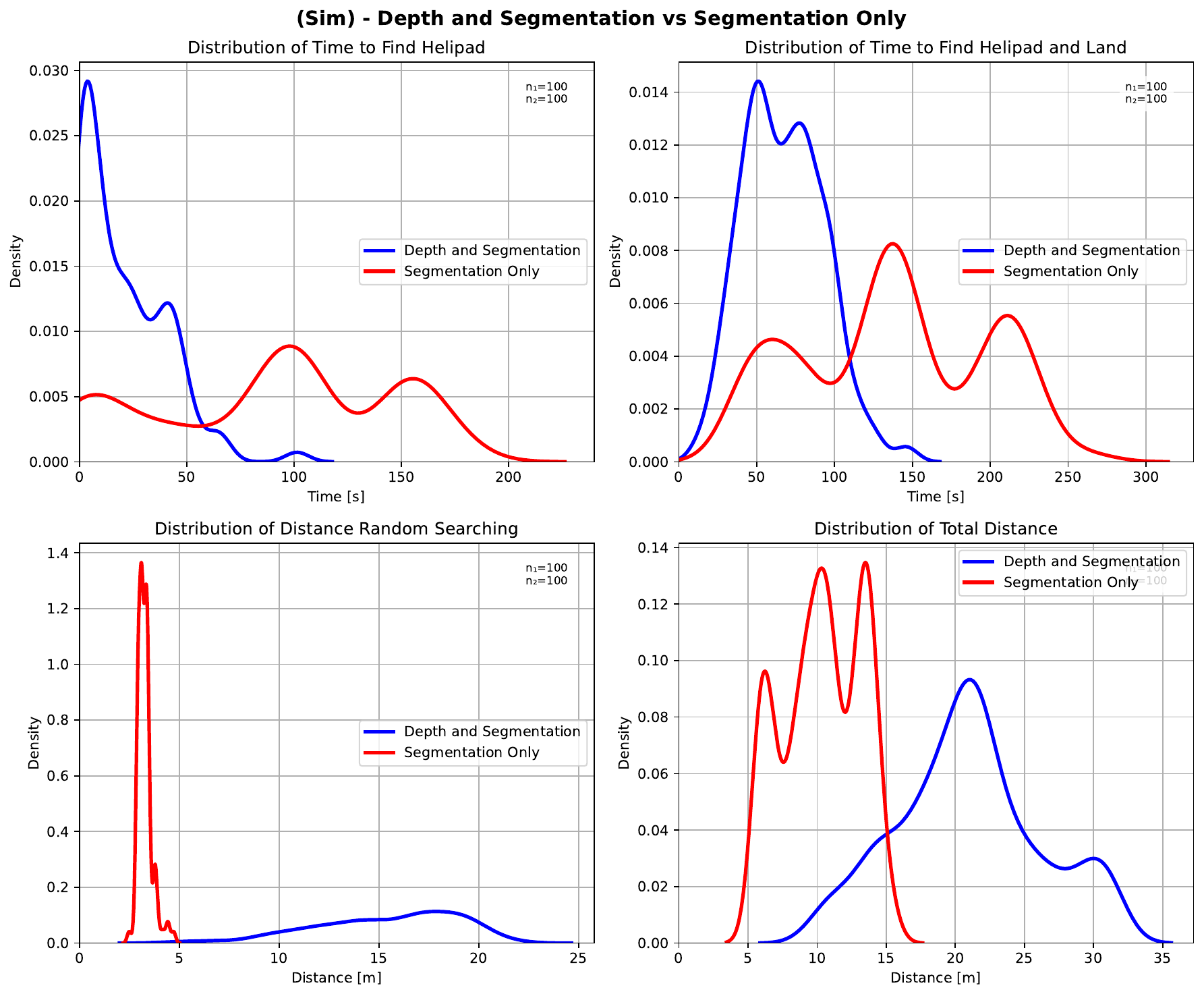}
        \caption{Comparison of the two methods used for the digital-twin environment. The top row represents the distribution of time to find helipad and the bottom one, the distance traveled.}
        \label{fig:kde_comparison_sim-depth-segmentation-tests-total_vs_sim-segmentation-tests-total}
    \end{minipage}
\end{figure*}

Based on the comparative analysis shown in Figure \ref{fig:kde_comparison_depth-segmentation-tests_vs_segmentation-tests} the integration of metric depth estimation with segmentation demonstrates substantial performance improvements over segmentation-only approach across multiple operational metrics. The safety corridor approach exhibits significantly faster convergence times for both helipad detection (distribution at around 10-15 seconds) and combined helipad landing task (peak density at approximately 20 seconds). The distance for random searching metric reveals a fundamental difference in search efficiency, while the segmentation-only method shows a distribution clustered at around 5 meters, indicating limited spatial exploration, the metric depth approach demonstrates a wider search pattern, extending to 25 meters. These results suggest that incorporating metric depth information enables for robust spatial exploring and navigation capabilities, leading to more efficient and reliable autonomous drone operations.

The results showed a strong correlation between real-world Figure \ref{fig:kde_comparison_depth-segmentation-tests_vs_segmentation-tests} and simulated Figure \ref{fig:kde_comparison_sim-depth-segmentation-tests-total_vs_sim-segmentation-tests-total} environments, validating our digital-twin approach. The depth and segmentation method consistently outperformed the segmentation-only approach in both distance efficiency and speed of finding the helipad.  

\subsection{Student Control Network}
\label{subsec:student-control-network}
To address the computational constraints of edge deployment, we developed a lightweight neural network ($1.6M$ parameters) that learns flight control policies from the vision-based autonomous flying teacher algorithms. The architecture was inspired by U-Net but modified for regression tasks by removing upsampling layers and replacing them with fully connected layers for direct movement prediction. A key advantage of this model is its compact size of just $1.6$ million parameters while still capturing knowledge from our previous best performing method. The network takes RGB frames as input with temporal context from two previous frames together with their corresponding labels. The 1 second temporal spacing for previous frames was selected through testing intervals ranging from $0.5s$ to $2s$. Testing was conducted using the same methodology described in Section \ref{subsec:vision-based-autonomous-flying}.

Initially, the drone achieved 30 successful flights out of 40 trials, with 10 crashes, the main reason for crashing was hitting an obstacle box. We first attempted to reduce the drone operational speed, hoping this would decrease crashes. However, after repeating the tests, there was no significant reduction in the crash rate. To address this challenge, we augmented the training dataset by duplicating data from flight segments near obstacles. This slightly improved performance, resulting in 35 successful flights out of 40 attempts.

\begin{figure}
    \centering
    \includegraphics[width=\linewidth]{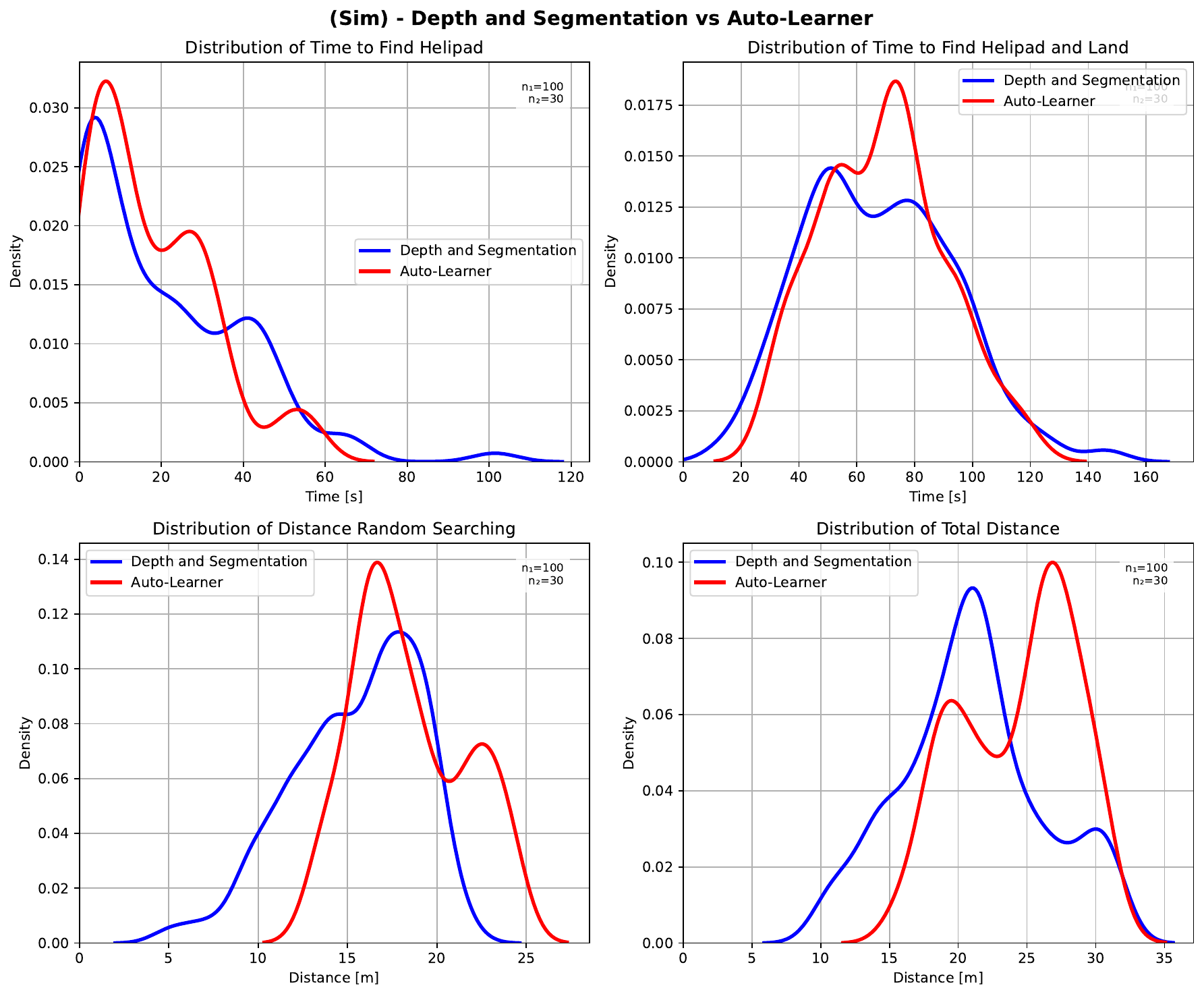}
    \caption{Comparison of the student network that learns from the best method in the simulated environment. The top row represents the distribution of time to find helipad and the bottom one, the distance traveled.}
    \label{fig:sim-comparison_depth-segmentation-tests_vs_auto-learner-tests}
\end{figure}

\section{Conclusion}
\label{sec:conclusion}
This paper demonstrates a vision-only autonomous flight system for small UAVs in controlled indoor environments, achieving reliable navigation through the integration of semantic segmentation and metric depth estimation. Our experimental validation across 130 flight tests establishes measurable performance improvements and identifies both the capabilities and limitations of current vision-based approaches.

The adaptive scale factor algorithm represents our primary technical contribution, successfully converting non-metric monocular depth predictions into metric measurements with a mean errors of 14.4 cm. This innovation enables real-time distance calculations essential for safe autonomous flight, addressing a fundamental challenge in vision-only navigation systems. The knowledge distillation framework, transitioning from a complex system having control algorithms and two neural networks to U-Net student (1.6M parameters), demonstrates that efficient semantic segmentation can be achieved on resource-constrained platforms while maintaining real-time performance.

We created a custom dataset in a controlled indoor environment that simulates urban landscapes, this provided resources for training and evaluating autonomous flight capabilities. Furthermore, the creation of the digital-twin in Unreal Engine and Parrot Sphinx simulator assured safe algorithm testing before releasing it in the real-world, significantly reducing the risk of hardware damage during development.

We then shown that training depth estimation and semantic segmentation models on domain-specific data substantially improves performance. Our approach for semantic segmentation, implemented through a knowledge distillation process from a SVM teacher to a U-Net student network, achieved real-time performance with high accuracy and low-effort labeling, in semantically segment elements crucial for the drone navigation.

The lightweight student control network, while showing promise for edge deployment with 87.5\% mission success rate, reveals the current limitations of end-to-end learning approaches. The 12.5\% failure rate, primarily due to collision events, indicates that direct policy learning from demonstration data requires more robust training strategies or additional safety mechanisms to match the reliability of the modular two-network approach.

Our current system operates efficiently in a controlled environment and may face challenges in more complex dynamic settings. However, to adapt to more complex environments one can replace the semantic segmentation model with state-of-the-art and achieve similar results. 

For the future, we would like to focus on improving a few aspects of our approaches. A first aspect would be improving the student control network to achieve comparable reliability of the two-model approach while maintaining computationally efficient. Future work will focus on extending the system capabilities to handle dynamic obstacles and implement more complex path planning algorithms that consider not only immediate obstacle avoidance but also global mission objectives.

\section*{Acknowledgment}
The authors would like to thank to the European Health and Digital Executive Agency (HADEA), under the powers delegated by the European Commission, through the DIGITWIN4CIUE project with grant agreement No. 101084054, Google Research Grant, "Romanian Hub for Artificial Intelligence - HRIA", Smart Growth, Digitization and Financial Instruments Program, 2021-2027 (MySMIS No. 334906) and "European Lighthouse of AI for Sustainability - ELIAS", Horizon Europe program (Grant No. 101120237).

\ifCLASSOPTIONcaptionsoff
  \newpage
\fi

\bibliographystyle{IEEEtran}
\bibliography{cas-refs}

\end{document}